%% file: main.tex
\documentclass{article}

\usepackage[utf8]{inputenc} 
\usepackage[T1]{fontenc}    
\usepackage{hyperref}       
\usepackage{url}            
\usepackage{booktabs}       
\usepackage{amsfonts}       
\usepackage{nicefrac}       
\usepackage{microtype}      

\input{custom}

\title{\LARGE \bf
Decentralized Deterministic Multi-Agent Reinforcement Learning 
}

\author{%
Antoine Grosnit$^{1}$, Desmond Cai$^{2}$, Laura Wynter$^{3}$
    \thanks{$^{1}$ Antoine Grosnit is from Ecole Polytechnique, Paris, France,
  \texttt{antoine.grosnit70@gmail.com}} \thanks{$^{2}$Desmond Cai is with AStar, Singapore
  \texttt{desmond.cai1@ibm.com}}
\thanks{$^{3}$Laura Wynter is with
  IBM Research, Singapore
  \texttt{lwynter@sg.ibm.com}}
}

\begin{document}

\maketitle

\begin{abstract}
  [Zhang, ICML 2018] provided
the first decentralized actor-critic algorithm for  multi-agent reinforcement learning (MARL) that offers convergence guarantees. 
In that work, policies are stochastic and are defined on finite action spaces. 
We extend those results to offer a provably-convergent decentralized actor-critic algorithm 
for learning deterministic policies on continuous action spaces. Deterministic policies are important in  real-world settings. To handle the lack of exploration inherent in deterministic policies, we consider both off-policy  and on-policy settings.
We provide  the expression of a local deterministic policy gradient, 
  decentralized deterministic actor-critic algorithms 
and  convergence guarantees for linearly-approximated  value functions. 
This work will help enable decentralized MARL in high-dimensional action spaces
and pave the way for more widespread use of MARL.
\end{abstract}

\input{introduction}
\input{background}

\input{gradient}

\input{algorithm}
\input{conclusion}

\input{supp}

\bibliographystyle{IEEEtran}
\bibliography{references}

\end{document}

%% file: custom.tex
\usepackage{amsmath}
\usepackage{amsfonts}
\usepackage{stmaryrd}
\usepackage{bbm}
\usepackage{amsthm}
\usepackage{multicol}
\usepackage{xcolor}
\usepackage{algorithmic}
\usepackage{algorithm}
\usepackage{subcaption}
\usepackage{graphicx}

\theoremstyle{plain}
\newtheorem{thm}{Theorem} 

\newtheorem{lem}{Lemma}

\theoremstyle{definition}
\newtheorem{asm}{Assumption} 
\newtheorem*{remark}{Remark} 
\newtheorem{cond}{Conditions}

\PassOptionsToPackage{svgnames}{color}

\newcommand{\pa}[1]{\left(#1\right)}
\newcommand{\pab}[1]{\left[#1\right]}
\newcommand{\paa}[1]{\left\{#1\right\}}

\newcommand{\pabb}[1]{\left\llbracket #1 \right\rrbracket}
\newcommand{\palV}[1]{\left\lVert #1 \right\lVert}
\newcommand{\pav}[1]{\left\vert #1 \right\vert}

\newcommand{\esp}[1]{\mathbb{E}\pab{#1}}

\newcommand{\espL}[2]{\mathbb{E}_{#2}\pab{#1}}

\newcommand{\nablaV}[3]{\left. \nabla_{#1} #2 \right|_{#1 = #3}}

\newcommand{\mc}[1]{\mathcal{#1}}

\newcommand{\limit}[1]{\underset{#1}{\text{ lim }}}

\usepackage{color}
\usepackage{ifthen}
\definecolor{darkgreen}{rgb}{0,0.5,0}
\newboolean{showcomments}
\setboolean{showcomments}{true}
\newcommand{\desmond}[1]{\ifthenelse{\boolean{showcomments}}{\textcolor{red}{(Desmond says: #1)}}{}}
\newcommand{\laura}[1]{\ifthenelse{\boolean{showcomments}}{\textcolor{red}{(Laura says: #1)}}{}}

%% file: introduction.tex
\section{Introduction}
\label{intro}

 Cooperative multi-agent reinforcement learning (MARL) has seen considerably less  use than its single-agent analog, in part because often no central agent exists to coordinate the  cooperative agents.  As a result, decentralized architectures have  been advocated for MARL. Recently, decentralized architectures have been shown to admit convergence guarantees comparable to their centralized counterparts under mild network-specific assumptions  (see \cite{FDMARL, Suttle:2019:MA-OFF-Policy-AC}).
In this work, we  develop a decentralized actor-critic algorithm with deterministic policies for  multi-agent reinforcement learning.  Specifically, we  extend  results
  for actor-critic  with stochastic policies  (\cite{Bhatnagar:2009:NaturalActorCritic, degris:2012:offPol_AC, Maei:2018:Convergent_AC_OffPol, Suttle:2019:MA-OFF-Policy-AC}) to handle  deterministic policies.
 Indeed,  theoretical and empirical work has shown that deterministic algorithms  outperform their stochastic counterparts in  high-dimensional continuous action settings  (\cite{Silver:2014:DeterministicPolicyGradient, Lillicrap:2015:DDPG, Fujimoto:2018:TD3}).
 Deterministic policies further avoid  estimating  the complex integral over the action space. Empirically this allows for lower variance of the critic estimates and faster convergence. On the other hand, deterministic policy gradient methods suffer from reduced exploration. For this reason, we provide both off-policy and on-policy versions of our results, the off-policy version allowing for significant improvements in exploration.
The  contributions of this paper are three-fold: (1) we derive the expression of the gradient in terms of the long-term average reward, which is needed in the undiscounted multi-agent setting with deterministic policies; (2) we show that the deterministic policy gradient is the limiting case, as policy variance tends to zero, of the stochastic policy gradient; and  (3) we provide a decentralized deterministic multi-agent actor critic algorithm and prove its  convergence under  linear function approximation. 

%% file: background.tex
\section{Background}
Consider a system of $N$ agents denoted by $\mc{N} = [N]$
 in a  decentralized setting.
Agents  determine their decisions independently 
based on observations of their own rewards. 
Agents may however communicate  via 
a possibly time-varying communication network, 
characterized by an undirected graph $\mc{G}_t = (\mc{N},\mc{E}_t)$,
where $\mc{E}_t$ is the set of communication links 
connecting the agents at time $t\in \mathbb{N}$. 
The networked multi-agent MDP is thus characterized by a tuple
$(\mc{S}, \paa{\mc{A}^i}_{i\in\mc{N}}, P, \paa{R^i}_{i\in\mc{N}}, \paa{\mc{G}_t}_{t\geq 0})$ where 
$\mc{S}$ is a finite global state space shared by all agents in $\mc{N}$,
$\mc{A}^i$ is the action space of agent $i$, and 
$\paa{\mc{G}_t}_{t\geq 0}$ is a time-varying communication network.
In addition, let $\mc{A} = \prod_{i\in\mc{N}} \mathcal{A}^i$ denote 
the joint action space of all agents.
Then, 
$P:\mc{S}\times\mc{A}\times\mc{S}\to[0,1]$ is the state transition probability of the MDP, and
$R^i:\mc{S}\times\mc{A}\to\mathbb{R}$ is the local reward function of agent $i$.
States and actions are assumed globally observable whereas rewards are only locally observable.
At time $t$, each agent $i$ chooses its action $a_t^i\in\mc{A}^i$ given state $s_t\in\mc{S}$, 
according to a local parameterized policy 
$\pi^i_{\theta^i}:\mc{S}\times\mc{A}^i\to[0,1]$,
where 
$\pi^i_{\theta^i}(s,a^i)$ 
is the probability of agent $i$ choosing action $a^i$ at state $s$, and 
$\theta^i\in\Theta^i \subseteq \mathbb{R}^{m_i}$ 
is the policy parameter.
We pack the parameters together as 
$\theta = [(\theta^1)^\top,\cdots,(\theta^N)^\top]^\top \in\Theta$ 
where 
$\Theta=\prod_{i\in\mc{N}}\Theta^i$. 
We denote the joint policy by 
$\pi_{\theta}:\mc{S}\times\mc{A}\to[0,1]$ 
where 
$\pi_{\theta}(s,a) = \prod_{i\in\mc{N}}\pi_{\theta^i}^i(s,a^i)$. 
Note that decisions are decentralized in  that 
rewards are observed locally, 
policies are evaluated locally, 
and actions are executed locally.
We assume that for any $i\in\mc{N}$, $s\in\mc{S}$, $a^i\in\mc{A}^i$, 
the policy function $\pi_{\theta^i}^i(s,a^i) > 0$ for any $\theta^i\in\Theta^i$
and that $\pi_{\theta^i}^i(s,a^i)$ is continuously differentiable with 
respect to the parameters $\theta^i$ over $\Theta^i$. 
In addition, for any $\theta\in\Theta$, 
let $P^\theta:\mc{S}\times\mc{S}\to[0,1]$ denote the transition matrix 
of the Markov chain $\paa{s_t}_{t\geq0}$ induced by policy $\pi_\theta$, 
that is, for any $s,s'\in\mc{S}$,
$
P^\theta(s'|s) = \sum_{a\in\mc{A}} \pi_\theta(s,a)\cdot P(s'|s,a).
$
We make the standard assumption that the Markov chain $\paa{s_t}_{t\geq0}$ 
is irreducible and aperiodic under any $\pi_\theta$ 
and denote its stationary distribution by $d_\theta$.

Our objective is to find a policy $\pi_\theta$ that 
maximizes the long-term average reward over the network. 
Let $r^i_{t+1}$ denote the reward received by agent $i$ as a result of taking action $a_t^i$. 
Then, we wish to solve:
\begin{align*}
  \max_\theta J(\pi_\theta)
  &= 
  \lim_{T\to\infty} \frac{1}{T} \mathbb{E} \pab{\sum_{t=0}^{T-1} \frac{1}{N}\sum_{i\in\mc{N}} r^i_{t+1}}
  = 
  \sum_{s\in\mc{S},a\in\mc{A}} d_\theta(s) \pi_\theta(s,a) \bar{R}(s,a),
\end{align*}
where $\bar{R}(s,a) = \pa{1/N}\cdot \sum_{i\in\mc{N}} R^i(s,a)$ 
is the globally averaged reward function. 
Let $\bar{r}_t = \pa{1/N}\cdot \sum_{i\in\mc{N}} r_t^i$, then
$\bar{R}(s,a) = \mathbb{E}\pab{\bar{r}_{t+1}|s_t=s,a_t=a}$, 
and therefore, the global relative action-value function is:
$
  Q_\theta(s,a)
  =$
$  \sum_{t\geq 0} \mathbb{E}\pab{\bar{r}_{t+1} - J(\theta) | s_0=s, a_0=a, \pi_\theta},
  $
and the global relative state-value function is:
$
  V_\theta(s) = \sum_{a\in\mc{A}} \pi_\theta(s,a) Q_\theta(s,a).
  $
For simplicity, we  refer to $V_\theta$ and $Q_\theta$ as simply the 
state-value function and action-value function.
We  define the advantage function as
$A_\theta(s,a) = Q_\theta(s,a) - V_\theta(s)$.


\cite{FDMARL} provided the first provably convergent 
MARL algorithm in the context of the above model.
The fundamental result underlying their algorithm is a
local policy gradient theorem:
\begin{equation*}
  \nabla_{\theta^i} J(\mu_\theta)
  =
  \mathbb{E}_{s\sim d_\theta,a\sim\pi_\theta}\pab{\nabla_{\theta^i} \log\pi^i_{\theta^i}(s,a^i) \cdot A^i_\theta(s,a)},
\end{equation*}
where 
$A_\theta^i(s,a) = Q_\theta(s,a) - \tilde{V}^i_\theta(s,a^{-i})$ 
is a local advantage function and 
$\tilde{V}^i_\theta(s,a^{-i})$ $= \sum_{a^i\in\mc{A}^i} \pi^i_{\theta^i}(s,a^i) Q_\theta(s,a^i,a^{-i})$.
This theorem has important practical value as it shows that 
the policy gradient with respect to each local parameter $\theta^i$
can be obtained locally using the corresponding score function $\nabla_{\theta^i}\log\pi^i_{\theta^i}$ 
provided that agent $i$ has an unbiased estimate of the advantage functions
$A^i_\theta$ or $A_\theta$. 
With only local information, the advantage functions $A^i_\theta$ or $A_\theta$ 
cannot be well estimated since the estimation 
requires the rewards $\paa{r_t^i}_{i\in\mc{N}}$ of all agents. 
Therefore, they proposed a consensus based actor-critic  
that leverages the communication network to share information between agents by placing a weight $c_t(i,j)$ on the message transmitted 
from agent $j$ to agent $i$ at time $t$. 
Their action-value function $Q_\theta$ was approximated by a 
parameterized function $\hat{Q}_{\omega}:\mc{S}\times\mc{A}\to\mathbb{R}$,
and each agent $i$ maintains its own parameter $\omega^i$,
which it uses to form a local estimate $\hat{Q}_{\omega^i}$ of the global $Q_\theta$.
At each time step $t$, each agent $i$ shares its local parameter $\omega_t^i$ 
with its neighbors on the network, and the shared parameters 
are used to arrive at a consensual estimate of $Q_\theta$ over time.


%% file: gradient.tex
\section{Local Gradients of Deterministic Policies}

While the use of a stochastic policy facilitates the derivations of  convergence proofs, most real-world control tasks require a deterministic policy to be implementable.
In addition, the quantities estimated in the deterministic critic  do not involve estimation of the complex integral over the action space found in the stochastic version. This offers lower variance of the critic estimates and  faster convergence. To address the lack of exploration that comes with deterministic policies, we provide both off-policy and on-policy versions of our results. Our first requirement is  a local deterministic policy gradient theorem. 

We assume that $\mc{A}^i = \mathbb{R}^{n_i}$. We make standard regularity assumptions on our MDP. That is, we assume that for any $s, s^\prime \in \mathcal{S}$, $P(s^\prime \vert s,a)$ and $R^i(s, a)$ are bounded and have bounded first and second derivatives.
We consider local deterministic policies $\mu^i_{\theta^i} : \mathcal{S} \to \mc{A}^i$ with parameter vector $\theta^i \in \Theta^i$, and denote the joint policy by $\mu_\theta:\mc{S}\to\mc{A}$, where $\mu_\theta(s)=(\mu_{\theta^1}^1(s),\ldots,\mu_{\theta^N}^N(s))$ and
$\theta=[(\theta^1)^\top,\ldots,(\theta^N)^\top]^\top$.
We assume that for any $s \in \mathcal{S}$, 
the deterministic policy function $\mu^i_{\theta^i}(s)$ is 
twice continuously differentiable with respect to 
the parameter $\theta^i$ over $\Theta^i$.
Let $P^\theta$ denote the transition matrix of 
the Markov chain $\{s_t\}_{t \geq 0}$ induced by policy $\mu_\theta$, 
that is, for any $s,s'\in\mc{S}$,
$
  P^{\theta}(s^\prime \vert s) = P(s^\prime \vert s, \mu_\theta(s)).
  $
We assume that the Markov chain $\{s_t\}_{t \geq 0}$ is 
irreducible and aperiodic under any $\mu_\theta$
and denote its stationary distribution by $d^{\mu_\theta}$.

Our objective is to find a policy $\mu_\theta$ that maximizes the long-run average reward:
\begin{align*}
  \max_\theta J(\mu_\theta)
  = 
  \mathbb{E}_{s\sim d^{\mu_\theta}} [\bar{R}(s,\mu_\theta(s))]
  =
  \sum_{s\in\mc{S}} d^{\mu_\theta}(s) \bar{R}(s,\mu_\theta(s)).
\end{align*}
Analogous to the stochastic policy case, we denote the action-value function by $Q_\theta(s,a) = \sum_{t\geq 0} \mathbb{E}[\bar{r}_{t+1} - J(\mu_\theta) | s_0 = s, a_0 = a, \mu_\theta]$, and the state-value function by $V_\theta(s) = Q_\theta(s,\mu_\theta(s))$. When there is no ambiguity, we will denote $J(\mu_\theta)$ and $d^{\mu_\theta}$ by simply $J(\theta)$ and $d^\theta$, respectively.
We present three results for the long-run average reward: (1) an expression for the local deterministic policy gradient in the on-policy setting $\nabla_{\theta^i} J(\mu_\theta)$, (2) an expression for the gradient in the off-policy setting, and (3) we show that the deterministic policy gradient can be seen as the limit of the stochastic one.



 \paragraph{On-Policy Setting}
 

\begin{thm}[Local Deterministic Policy Gradient Theorem - On Policy] \label{thm:On_Policy_Grad} 
For any $\theta \in \Theta$, $i \in \mathcal{N}$, 
$\nabla_{\theta^i} J(\mu_\theta)$ exists and is given by
\begin{equation*}
    \nabla_{\theta^i}J(\mu_\theta) = 
    \espL{\nabla_{\theta^i} \mu^i_{\theta^i}(s) \nabla_{a^i} \left. Q_\theta(s, \mu^{-i}_{\theta^{-i}}(s), a^i)\right|_{a^i = \mu^i_{\theta^i}(s)}}{s \sim d^{\mu_\theta}}.
\end{equation*}
\end{thm}
The first step of the proof consists in showing that $\nabla_\theta J(\mu_\theta) = \espL{\nabla_\theta \mu_\theta(s) \nabla_a \left. Q_\theta(s,a)\right|_{a = \mu_\theta(s)}}{s \sim d^\theta}$. This is an extension of the well-known stochastic case, for which we have $\nabla_\theta J(\pi_\theta) = \espL{\nabla_\theta \log(\pi_\theta(a|s)) Q_\theta(s,a)}{s \sim d_\theta}$, which holds for a long-term averaged return with stochastic policy (e.g Theorem 1 of \cite{Policy_grad_meth_RL_fct_appr}). See the Appendix for the details.

  \paragraph{Off-Policy Setting}

In the off-policy setting, we are given a behavior policy $\pi:\mc{S}\to\mc{P}(\mc{A})$, and our goal is to maximize the long-run average reward under state distribution $d^{\pi}$:
\begin{equation} \label{eq:off_policy_objective_fct}
    J_\pi(\mu_\theta) = \mathbb{E}_{s \sim d^\pi} \big[\bar{R}(s,\mu_\theta(s))\big] = \sum_{s \in \mathcal{S}} d^\pi(s) \bar{R}(s,\mu_\theta(s)).
\end{equation}

Note that we consider here an excursion objective (\cite{Sutton:2009:FGM:excursion, DeterministicPolicyGradient, Sutton:2016:EAP:excursion}) since we take the average over the state distribution of the behaviour policy $\pi$ of the state-action reward when selecting action given by the target policy $\mu_\theta$. We thus have:

\begin{thm}[Local Deterministic Policy Gradient Theorem - Off Policy] \label{thm:Off_Policy_Grad} 
For any $\theta \in \Theta$, $i \in \mathcal{N}$, $\pi: \mc{S} \to \mc{P}(\mc{A})$ a fixed stochastic policy, 
$\nabla_{\theta^i} J_\pi(\mu_\theta)$ exists and is given by
\begin{equation*}
    \nabla_{\theta^i} J_\pi(\mu_\theta) = 
    \espL{\nabla_{\theta^i} \mu^i_{\theta^i}(s) \nabla_{a^i} \left. \bar{R}(s, \mu^{-i}_{\theta^{-i}}(s), a^i)\right|_{a^i = \mu^i_{\theta^i}(s)}}{s \sim d^\pi}.
\end{equation*}
\end{thm}

\begin{proof}
Since $d^\pi$ is independent of $\theta$ we can take the gradient on both sides of (\ref{eq:off_policy_objective_fct})
\begin{equation*}
    \nabla_\theta J_\pi(\mu_\theta) = \espL{\nabla_\theta \mu_\theta(s) \left. \nabla_a \bar{R}(s,\mu_\theta(s)) \right|_{a = \mu_\theta(s)}}{s \sim d^\pi}.
\end{equation*}
Given that $\nabla_{\theta^i} \mu_{\theta}^j (s) = 0$ if $i \neq j$, we have $\nabla_\theta \mu_\theta(s) = \text{Diag}(\nabla_{\theta^1} \mu_{\theta_1}^1 (s), \dots, \nabla_{\theta^N} \mu_{\theta_N}^N (s))$ and the result follows.
\end{proof}
This result implies that,  off-policy,  each agent needs access  to $\mu_{\theta_t^{-i}}^{-i}(s_t)$ for every $t$ .

\paragraph{Limit Theorem}

As noted by \cite{Silver:2014:DeterministicPolicyGradient},  the fact that the deterministic gradient is a limit case of the stochastic gradient enables the  standard machinery of policy gradient, such as compatible-function approximation (\cite{Sutton:2000:Policy_grad_meth_RL_fct_appr}), natural gradients (\cite{Kakade:2001:NPG}), on-line feature adaptation (\cite{Prabuchandran:2016:A-C_Online_Feature_Adapt},) and  actor-critic (\cite{Konda:2002:Actor-Critic}) to be used with deterministic policies. 
We show that it holds in our setting.  The proof can be found in the Appendix.
\begin{thm}[Limit of the Stochastic Policy Gradient for MARL] \label{thm:limit_stoch_grad}
Let $\pi_{\theta, \sigma}$ be a stochastic policy such that $\pi_{\theta, \sigma}(a|s) = \nu_\sigma(\mu_\theta(s),a)$, where $\sigma$ is a parameter controlling the variance, and $\nu_\sigma$ satisfy Condition~\ref{cond:regular_delta_appr} in the Appendix. 
Then,
\begin{equation*}
     \underset{\sigma \downarrow 0}{\textnormal{lim }} \nabla_\theta J_{\pi_{\theta, \sigma}} (\pi_{\theta, \sigma}) = \nabla_\theta J_{\mu_\theta}(\mu_\theta)
\end{equation*}
where on the l.h.s the gradient is the standard stochastic policy gradient and on the r.h.s. the gradient is the deterministic policy gradient.
\end{thm}

%% file: algorithm.tex
\section{Algorithms}

We provide  two decentralized deterministic actor-critic algorithms, one on-policy and the other off-policy and demonstrate their convergence in the next section; assumptions and proofs are provided in the Appendix.

\paragraph{On-Policy Deterministic Actor-Critic}
\begin{algorithmic}
\begin{algorithm}[ht]
   \STATE Initialize $\hat{J}_0^i, \omega^i_0, \widetilde{\omega}^i_0, \theta_0^i, \forall i \in \mathcal{N}$; state $s_0$; stepsizes $\{\beta_{\omega, t}\}_{t \geq 0}, \{\beta_{\theta, t}\}_{t \geq 0}$
   \STATE Draw $a^i_0 = \mu^i_{\theta^i_0}(s_0)$ \textcolor{blue}{and compute $\widetilde{a}_0^i = \nabla_{\theta^i} \mu^i_{\theta^i_0}(s_0)$}
    \STATE Observe joint action $a_0 = (a^1_0, \dots, a^N_0)$ \textcolor{blue}{and $\widetilde{a}_0 = \left(\widetilde{a}^1_0, \dots, \widetilde{a}^N_0\right)$}
   \REPEAT
     \FOR{$i \in \mathcal{N}$}
    \STATE Observe $s_{t+1}$ and reward $r_{t+1}^i = r^i(s_t, a_t)$
    \STATE Update $\hat{J}^i_{t+1} \leftarrow (1 - \beta_{\omega, t}) \cdot \hat{J}_t^i + \beta_{\omega, t} \cdot r_{t+1}^i$
    \STATE Draw action $a_{t+1} = \mu^i_{\theta^i_t} (s_{t+1})$ \textcolor{blue}{and compute $\widetilde{a}_{t+1}^i = \nabla_{\theta^i} \mu^i_{\theta^i_t}(s_{t+1})$}
   \ENDFOR
   \STATE Observe joint action $a_{t+1} = (a^1_{t+1}, \dots, a^N_{t+1})$ \textcolor{blue}{and $\widetilde{a}_{t+1} = \left(\widetilde{a}^1_{t+1}, \dots, \widetilde{a}^N_{t+1}\right)$}
   \FOR{$i \in \mathcal{N}$}
   \STATE Update: $\delta_t^i \leftarrow r_{t+1}^i - \hat{J}_t^i + \hat{Q}_{\omega_t^i} (s_{t+1}, a_{t+1}) - \hat{Q}_{\omega_t^i}(s_t, a_t)$
   \STATE \textbf{Critic step: } $\widetilde{\omega}^i_t \leftarrow \omega_t^i + \beta_{\omega, t} \cdot \delta_t^i \cdot \left. \nabla_\omega \hat{Q}_{\omega^i} (s_t,a_t) \right|_{\omega = \omega^i_t}$
   \STATE \textbf{Actor step: } $\theta^i_{t+1} = \theta_t^i + \beta_{\theta, t} \cdot \nabla_{\theta^i} \mu_{\theta^i_t}^i(s_t) \left. \nabla_{a^i} \hat{Q}_{\omega^i_t} (s_t,a^{-i}_t, a^i)\right|_{a^i = a^i_t}$
   \STATE Send $\widetilde{\omega}^i_t$ to the neighbors $\{j \in \mathcal{N} : (i,j) \in \mathcal{E}_t \}$ over $\mathcal{G}_t$
   \STATE \textbf{Consensus step: } $\omega^i_{t+1} \leftarrow \sum_{j \in \mathcal{N}} c^{i j}_t \cdot \widetilde{\omega}^j_t$
   \ENDFOR
   \UNTIL{end}
   \caption{Networked deterministic on-policy actor-critic}
   \label{alg:on_policy}
\end{algorithm}
\end{algorithmic}
Consider the following on-policy algorithm. 
The actor step   is based on an expression for $\nabla_{\theta^i}J(\mu_\theta)$ in terms of $\nabla_{a^i} Q_\theta$(see Equation  \eqref{local_grad_J_Q} in the Appendix). We approximate the action-value function $Q_\theta$ using a family of functions $\hat{Q}_\omega : \mathcal{S} \times \mathcal{A} \to \mathbb{R}$ parameterized by $\omega$, a column vector in $\mathbb{R}^K$. Each agent $i$ maintains its own parameter $\omega^i$ and  uses $\hat{Q}_{\omega^i}$ as its local estimate of $Q_\theta$. The parameters $\omega^i$ are updated in the critic step using consensus updates  through a weight matrix $C_t =\pa{c^{i j}_t}_{i,j} \in \mathbb{R}^{N\times N}$ where $c_t^{i j}$ is the weight on the message transmitted from $i$ to $j$ at time $t$, namely:
\begin{align} \label{alg:On_Policy_AC_critic_step}
&\hat{J}_{t+1}^i = (1 - \beta_{\omega, t}) \cdot \hat{J}_t^i + \beta_{\omega, t} \cdot r_{t+1}^i \\
&\widetilde{\omega}^i_t = \omega_t^i + \beta_{\omega, t} \cdot \delta_t^i \cdot \left. \nabla_\omega \hat{Q}_{\omega^i} (s_t,a_t) \right|_{\omega = \omega^i_t} \\
&\omega_{t+1}^i = \sum_{j \in \mathcal{N}} c_t^{i j} \cdot \widetilde{\omega}^j_t
\end{align}
with
\begin{equation*}
    \delta_t^i = r_{t+1}^i - \hat{J}_t^i + \hat{Q}_{\omega_t^i} (s_{t+1}, a_{t+1}) - \hat{Q}_{\omega_t^i}(s_t, a_t).
\end{equation*}
For the actor step, each agent $i$ improves its policy via:
\begin{equation} \label{algo1_actor_step}
    \theta^i_{t+1} = \theta_t^i + \beta_{\theta, t} \cdot \nabla_{\theta^i} \mu_{\theta^i_t}^i(s_t) \cdot \left. \nabla_{a^i} \hat{Q}_{\omega^i_t} (s_t,a^{-i}_t, a^i)\right|_{a^i = a^i_t}.
\end{equation}
Since Algorithm \ref{alg:on_policy} is an on-policy algorithm, each agent updates the critic using only $(s_t, a_t, s_{t+1})$, at time $t$ knowing that $a_{t+1} = \mu_{\theta_t}(s_{t+1})$.
The terms in blue are additional terms that need to be shared when using compatible features (this is explained further in the next section).


\paragraph{Off-Policy Deterministic Actor-Critic}

We  further propose an  off-policy actor-critic algorithm, defined in Algorithm \ref{alg:Off_Policy} to enable better exploration capability.  Here, the goal is to maximize $J_\pi(\mu_\theta)$ where $\pi$ is the behavior policy. To do so, the globally averaged reward function $\bar{R}(s, a)$ is approximated using a family of functions $\hat{\bar{R}}_\lambda: \mathcal{S} \times \mathcal{A} \to \mathbb{R}$ that are parameterized by $\lambda$, a column vector in $\mathbb{R}^K$. Each agent $i$ maintains its own parameter $\lambda^i$ and  uses $\hat{\bar{R}}_{\lambda^i}$ as its local estimate of $\bar{R}$. 
Based on (\ref{thm:On_Policy_Grad}), the actor update is
\begin{equation} \label{algo2_actor_step}
    \theta^i_{t+1} = \theta^i_t + \beta_{\theta, t} \cdot \nabla_{\theta^i} \mu_{\theta^i_t}^i(s_t) \cdot \left. \nabla_{a^i} \hat{\bar{R}}_{\lambda^i_t} (s_t,\mu_{\theta_t^{-i}}^{-i}(s_t), a^i)\right|_{a^i = \mu_{\theta^i_t}(s_t)},
\end{equation}
which requires each agent $i$ to have access to $\mu_{\theta_t^j}^j(s_t)$ for $j \in \mathcal{N}$.
\bigbreak
The critic update is
\begin{align}\label{alg:Off_Policy_AC_critic_step}
    &\widetilde{\lambda}^i_t = \lambda_t^i + \beta_{\lambda, t} \cdot \delta_t^i \cdot \left. \nabla_\lambda \hat{\bar{R}}_{\lambda^i} (s_t,a_t)\right|_{\lambda = \lambda^i_t} \\
    &\lambda_{t+1}^i = \sum_{j \in \mathcal{N}} c_t^{i j} \widetilde{\lambda}^j_t, 
\end{align}
with
\begin{equation}
    \delta_t^i = r^i(s_t, a_t) - \hat{\bar{R}}_{\lambda^i_t}(s_t, a_t).
\end{equation}
In this case, $\delta_t^i$ was motivated by distributed optimization results, and  is not related to the local TD-error (as there is no "temporal" relationship for $R$). Rather, it is simply the difference between the sample reward and the bootstrap estimate. The terms in blue are additional terms that need to be shared when using compatible features (this is explained further in the next section).
\begin{algorithm}[htb]
   \caption{Networked deterministic off-policy actor-critic}
   \label{alg:Off_Policy}
\begin{algorithmic}
 \STATE Initialize $\lambda^i_0, \widetilde{\lambda}^i_0, \theta_0^i, \forall i \in \mathcal{N}$; state $s_0$; stepsizes $\{\beta_{\lambda, t}\}_{t \geq 0}, \{\beta_{\theta, t}\}_{t \geq 0}$
 \STATE Draw $a^i_0 \sim \pi^i(s_0)$ , compute $\dot{a}^i_0 = \mu^i_{\theta^i_0}(s_0)$ \textcolor{blue}{and $\widetilde{a}_0^i = \nabla_{\theta^i} \mu^i_{\theta^i_0}(s_0)$}
 \STATE Observe joint action $a_0 = (a^1_0, \dots, a^N_0)$, $\dot{a}_0 = (\dot{a}^1_0, \dots, \dot{a}^N_0)$ \textcolor{blue}{and $\widetilde{a}_0 = \pa{\widetilde{a}^1_0, \dots, \widetilde{a}^N_0}$}
\REPEAT
  \FOR{$i \in \mathcal{N}$}
  \STATE Observe $s_{t+1}$ and reward $r^i_{t+1} = r^i(s_t, a_t)$
  \ENDFOR
  \FOR{$i \in \mathcal{N}$}
  \STATE Update: $\delta_t^i \leftarrow r_{t+1}^i - \hat{\bar{R}}_{\lambda^i_t}(s_t, a_t)$
  \STATE \textbf{Critic step: } $\widetilde{\lambda}^i_t \leftarrow \lambda_t^i + \beta_{\lambda, t} \cdot \delta_t^i \cdot \left. \nabla_\lambda \hat{\bar{R}}_{\lambda^i} (s_t,a_t)\right|_{\lambda = \lambda^i_t}$
  \STATE \textbf{Actor step: } $\theta^i_{t+1} = \theta^i_t + \beta_{\theta, t} \cdot \nabla_{\theta^i} \mu_{\theta^i_t}^i(s_t) \cdot \left. \nabla_{a^i} \hat{\bar{R}}_{\lambda^i_t} (s_t,\mu_{\theta^{-i}_t}^{-i}(s_t), a^i)\right|_{a^i = \mu_{\theta^i_t}(s_t)}$
  \STATE Send $\widetilde{\lambda}^i_t$ to the neighbors $\{j \in \mathcal{N} : (i,j) \in \mathcal{E}_t \}$ over $\mathcal{G}_t$
  \ENDFOR
  \FOR{$i \in \mathcal{N}$}
  \STATE \textbf{Consensus step: } $\lambda^i_{t+1} \leftarrow \sum_{j \in \mathcal{N}} c^{i j}_t \cdot \widetilde{\lambda}^j_t$
  \STATE Draw action $a_{t+1} \sim \pi(s_{t+1})$, compute $\dot{a}^i_{t+1} = \mu^i_{\theta^i_{t+1}}(s_{t+1})$ \textcolor{blue}{and compute $\widetilde{a}_{t+1}^i = \nabla_{\theta^i} \mu^i_{\theta^i_{t+1}}(s_{t+1})$}
  \ENDFOR
  \STATE Observe joint action $a_{t+1} = (a^1_{t+1}, \dots, a^N_{t+1})$, $\dot{a}_{t+1} = (\dot{a}^1_{t+1}, \dots, \dot{a}^N_{t+1})$ \textcolor{blue}{and $\widetilde{a}_{t+1} = \pa{\widetilde{a}^1_{t+1}, \dots, \widetilde{a}^N_{t+1}}$}\\
  \UNTIL{end}
\end{algorithmic}
\end{algorithm}

\section{Convergence }

To show convergence, we use a two-timescale technique where in the actor, updating deterministic policy parameter $\theta^i$ occurs more slowly than that of  $\omega^i$ and $\hat{J}^i$ in the critic. We  study the asymptotic behaviour of the critic  by freezing the joint policy $\mu_\theta$,  then  study the behaviour of $\theta_t$ under  convergence of the critic.  
To ensure  stability, projection is often assumed since  it is not clear how boundedness of $\paa{\theta_t^i}$ can  otherwise be  ensured (see \cite{Bhatnagar:2009:NaturalActorCritic}). However, in practice, convergence is typically observed even without the projection step (see \cite{Bhatnagar:2009:NaturalActorCritic, degris:2012:offPol_AC, Prabuchandran:2016:A-C_Online_Feature_Adapt, FDMARL, Suttle:2019:MA-OFF-Policy-AC}). Additional technical  assumptions are required to show convergence and can be found in the Appendix.


\paragraph{On-Policy  Convergence}

To state convergence of the critic step, we define $D^s_\theta = \text{Diag}\big[d^\theta(s), s \in \mathcal{S}\big]$, $\bar{R}_\theta = \big[\bar{R}(s, \mu_\theta(s)), s \in \mathcal{S}\big]^\top \in \mathbb{R}^{\vert \mathcal{S} \vert}$ and the operator $T^Q_\theta : \mathbb{R}^{\vert \mathcal{S} \vert} \to \mathbb{R}^{\vert \mathcal{S} \vert}$ for any action-value vector $Q \in \mathbb{R}^{\vert \mathcal{S} \vert}$ (and not $\mathbb{R}^{\vert \mathcal{S} \vert \cdot \vert \mathcal{A} \vert}$ since there is a mapping associating an action to each state) as:  
$$T_\theta^Q(Q') = \bar{R}_\theta - J(\mu_\theta) \cdot \textbf{1} + P^\theta Q'.$$
\begin{thm}\label{thm:on-policy-AC-critic-cv}
Under Assumptions 
 \ref{Bound_theta}, \ref{asm_random_matrix}, and 
\ref{Two_timesteps}, for any given deterministic policy $\mu_\theta$, with $\{\hat{J}_t\}$ and $\{\omega_t\}$ generated from (\ref{alg:On_Policy_AC_critic_step}), we have $\textnormal{lim}_{t \to \infty} \frac{1}{N}\sum_{i \in \mathcal{N}} \hat{J}_t^i = J(\mu_\theta)$ and $\textnormal{lim}_{t \to \infty} \omega_t^i = \omega_\theta$ a.s. for any $i \in \mathcal{N}$, where
$$J(\mu_\theta) = \sum_{s \in \mathcal{S}} d^\theta(s) \bar{R}(s, \mu_\theta(s))$$
is the long-term average return under $\mu_\theta$, and $\omega_\theta$ is the unique solution to
\begin{equation} \label{omega_expr}
{\Phi_\theta}^\top D_\theta^s \big[T_\theta^Q(\Phi_\theta \omega_\theta) - \Phi_\theta \omega_\theta \big] = 0.
\end{equation}
Moreover, $\omega_\theta$ is the minimizer of the Mean Square Projected Bellman Error (MSPBE), i.e., the solution to
$$\underset{\omega}{\textnormal{minimize }} \lVert \Phi_\theta \omega - \Pi T_\theta^Q(\Phi_\theta \omega) \lVert^2_{D_\theta^s},$$
where $\Pi$ is the operator that projects a vector to the space spanned by the columns of $\Phi_\theta$, and $\lVert \cdot \lVert^2_{D_\theta^s}$ denotes the euclidean norm weighted by the matrix $D_\theta^s$.
\end{thm}
To state convergence of the actor step, we define quantities $\psi_{t, \theta}^i$, $\xi_t^i$ and $\xi_{t, \theta}^i$ as
\begin{align*}
&\psi_{t, \theta}^i = \nabla_{\theta^i} \mu_{\theta^i}^i(s_t) \quad \text{and} \quad
\psi_{t}^i = \psi_{t, \theta_t}^i = \nabla_{\theta^i} \mu_{\theta^i_t}^i(s_t),
\\
& \xi_{t, \theta}^i = \nablaV{a_i}{\hat{Q}_{\omega_\theta} (s_t, a_t^{-i}, a_i)}{a_i = \mu^i_{\theta^i_t}(s_t)} = \nablaV{a_i}{\phi(s_t, a_t^{-i}, a_i)}{a_i = \mu^i_{\theta^i_t}(s_t)} \omega_\theta,
\\
& \xi_t^i = \left. \nabla_{a_i} \hat{Q}_{\omega^i_t} (s_t, a_t^{-i}, a_i) \right|_{a_i = \mu^i_{\theta^i}(s_t)} = \left. \nabla_{a_i} \phi(s_t, a_t^{-i}, a_i) \right|_{a_i = \mu^i_{\theta^i}(s_t)} \omega_t^i.
\end{align*}
Additionally, we introduce the operator $\hat{\Gamma}(\cdot)$ as
\begin{equation} \label{eq_projection_limit}
    \hat{\Gamma}^i\pab{g(\theta)} = \underset{0<\eta \to 0}\lim \frac{\Gamma^i\pab{\theta^i + \eta \cdot g(\theta)} - \theta^i}{\eta}
\end{equation}
for any $\theta \in \Theta$ and $g:\Theta \to \mathbb{R}^{m_i}$ a continuous function. In case the limit above is not unique we take $\hat{\Gamma}^i\pab{g(\theta)}$ to be the set of all possible limit points of (\ref{eq_projection_limit}).
 \begin{thm} \label{thm:on-policy-AC-actor-cv}
 Under Assumptions 
\ref{Linear_appr}, 
 \ref{Bound_theta}, 
\ref{asm_random_matrix}, and 
\ref{Two_timesteps}, the policy parameter $\theta_t^i$ obtained from (\ref{algo1_actor_step}) converges a.s. to a point in the set of asymptotically stable equilibria of
 \begin{equation} \label{ODE_actor}
 \dot{\theta}^i = \hat{\Gamma}^i\pab{\espL{\psi_{t,\theta}^i \cdot \xi_{t, \theta}^i}{s_t \sim d^\theta, \mu_\theta}}, \quad \textnormal{for any } i \in \mathcal{N}.
 \end{equation}
 In the case of multiple limit points, the above is treated as a differential inclusion rather than an ODE.
 \end{thm}
The convergence of the critic step can be proved by taking similar steps as that in~\cite{FDMARL}. For the convergence of the actor step, difficulties arise from the projection (which is handled using Kushner-Clark Lemma~\cite{Kushner_Clark}) and the state-dependent noise (that is handled by ``natural'' timescale averaging~\cite{BorkarStochasticApprDynamicalSysViewpoint}). Details are provided in the Appendix.
\begin{remark} \label{rk_algo1_compatible_features}
Note that that with a linear function approximator  $Q_\theta$, $\psi_{t, \theta} \cdot \xi_{t,\theta} = \nabla_\theta \mu_\theta(s_t) \left. \nabla_{a} \hat{Q}_{\omega_\theta}(s_t,a)\right|_{a = \mu_\theta(s_t)}$ may not be an unbiased estimate of $\nabla_{\theta} J(\theta)$:
\begin{equation*}
    \mathbb{E}_{s \sim d^\theta}\big[\psi_{t, \theta} \cdot \xi_{t, \theta} \big] = \nabla_\theta J(\theta) + \espL{\nabla_\theta \mu_\theta(s) \cdot \pa{\nablaV{a}{\hat{Q}_{\omega_\theta}(s,a)}{\mu_\theta(s)} - \nablaV{a}{Q_{\omega_\theta}(s,a)}{\mu_\theta(s)}}}{s \sim d^\theta}.
\end{equation*}
A standard approach to overcome this approximation issue is via compatible features (see, for example, \cite{DeterministicPolicyGradient} and \cite{DistributedOffPolicyActorCriticRL}), i.e. $\phi(s,a) = a \cdot \nabla_\theta \mu_\theta(s)^\top$, giving, for $\omega \in \mathbb{R}^m$,
\begin{align*}
    &\hat{Q}_{\omega}(s,a) 
    = a \cdot \nabla_\theta \mu_\theta(s)^\top \omega 
    = (a - \mu_\theta(s))  \cdot \nabla_\theta \mu_\theta(s)^\top \omega + \hat{V}_{\omega}(s), 
    \\
    &\qquad\text{with } \hat{V}_{\omega}(s) 
    = \hat{Q}_{\omega}(s,\mu_\theta(s))
    \;\; \text{and  } \;\; \left. \nabla_{a} \hat{Q}_{\omega}(s,a) \right|_{a = \mu_{\theta}(s)} = \nabla_\theta \mu_\theta(s)^\top \omega.
\end{align*}
We thus expect that the  convergent point of \eqref{thm:on-policy-AC-actor-cv} corresponds to a small neighborhood of a local optimum of $J(\mu_\theta)$, i.e., $\nabla_{\theta^i} J(\mu_\theta) = 0$, provided that the error for the gradient of the action-value function $\left. \nabla_{a} \hat{Q}_{\omega}(s,a) \right|_{a = \mu_{\theta}(s)} - \left. \nabla_{a} Q_\theta(s,a)\right|_{a = \mu_\theta(s)}$ is small. However, note that using compatible features requires computing, at each step $t$, $\phi(s_t, a_t) = a_t \cdot \nabla_\theta \mu_\theta(s_t)^\top$. Thus, in Algorithm \ref{alg:on_policy}, each agent observes not only the joint action $a_{t+1} = (a_{t+1}^1, \dots, a_{t+1}^N)$ but also $(\nabla_{\theta^1} \mu^1_{\theta^1_t}(s_{t+1}), \dots, \nabla_{\theta^N} \mu^N_{\theta^N_t}(s_{t+1}))$ (see the parts in blue in Algorithm~\ref{alg:on_policy}).
\end{remark}

\paragraph{Off-Policy  Convergence}

\begin{thm} \label{thm:off-policy-AC-critic-cv}
Under Assumptions 
\ref{asm_compatible_features_R},
\ref{asm_random_matrix}, and   \ref{asm_algo2_two_timesteps},
for any given behavior policy $\pi$ and any $\theta \in \Theta$, with $\{\lambda_t^i\}$ generated from (\ref{alg:Off_Policy_AC_critic_step}), we have $\textnormal{lim}_{t \to \infty} \lambda^i_t = \lambda_\theta$ a.s. for any $i \in \mathcal{N}$, where $\lambda_\theta$ is the unique solution to
\begin{equation} \label{eq_algo2_consensus_cv}
    B_{\pi, \theta}\cdot\lambda_\theta = A_{\pi,\theta} \cdot d_\pi^s
\end{equation}
where $d_\pi^s = \big[d^\pi(s), s \in \mathcal{S}\big]^\top$, $A_{\pi,\theta} = \big[\int_\mathcal{A}\pi(a|s) \bar{R}(s,a) w(s,a)^\top \textnormal{d}a, s \in \mathcal{S}\big] \in \mathbb{R}^{K \times \vert \mathcal{S} \vert}$ and $B_{\pi, \theta} = \big[\sum_{s\in\mathcal{S}} d^\pi(s) \int_\mathcal{A}\pi(a|s) w_i(s,a) \cdot w(s,a)^\top \textnormal{d}a, 1 \leq i \leq K\big] \in \mathbb{R}^{K \times K}$.
\end{thm}
From here on we let
\begin{align*}
& \xi_{t, \theta}^i = \left. \nabla_{a_i} \hat{\Bar{R}}_{\lambda_\theta} (s_t,  \mu_{\theta_t^{-i}}^{-i} (s_t), a_i) \right|_{a_i = \mu^i_{\theta^i_t}(s_t)} = \left. \nabla_{a_i} w(s_t,  \mu_{\theta_t^{-i}}^{-i} (s_t), a_i) \right|_{a_i = \mu^i_{\theta^i_t}(s_t)} \lambda_\theta  \\
& \xi_t^i = \left. \nabla_{a_i} \hat{\bar{R}}_{\lambda^i_t} (s_t,  \mu_{\theta_t^{-i}}^{-i} (s_t), a_i) \right|_{a_i = \mu^i_{\theta^i_t}(s_t)} = \left. \nabla_{a_i} w(s_t,  \mu_{\theta^{-i}}^{-i} (s_t), a_i) \right|_{a_i = \mu^i_{\theta^i}(s_t)} \lambda_t^i
\end{align*}
and we keep
\begin{align*}
    \psi_{t, \theta}^i = \nabla_{\theta^i} \mu_{\theta^i}^i(s_t), \quad \text{and} \quad
    \psi_{t}^i = \psi_{t, \theta_t}^i = \nabla_{\theta^i} \mu_{\theta^i_t}^i(s_t).
\end{align*}
\begin{thm} \label{thm:off-policy-AC-actor-cv}
Under Assumptions 
\ref{asm_compatible_features_R},
\ref{Bound_theta}, 
\ref{asm_random_matrix}, and
 \ref{asm_algo2_two_timesteps},
the policy parameter $\theta_t^i$ obtained from (\ref{algo2_actor_step}) converges a.s. to a point in the asymptotically stable equilibria of
\begin{equation}
    \dot{\theta}^i = \Gamma^i\pab{\espL{\psi_{t,\theta}^i \cdot \xi_{t,\theta}^i}{s \sim d^\pi}}.
\end{equation}
\end{thm}
We define compatible features for the action-value and the average-reward  function in an analogous manner: $w_\theta(s,a) = (a - \mu_\theta(s)) \cdot \nabla_\theta \mu_\theta(s)^\top$. For $\lambda \in \mathbb{R}^m$,
\begin{align*}
    &\hat{\bar{R}}_{\lambda, \theta}(s,a) = (a - \mu_\theta(s)) \cdot \nabla_\theta \mu_\theta(s)^\top \cdot \lambda \\
    &\nabla_a \hat{\bar{R}}_{\lambda, \theta} (s,a) = \nabla_\theta \mu_\theta(s)^\top \cdot \lambda
\end{align*}
and we have that, for $\lambda^* = \underset{\lambda}{\textnormal{argmin }} \mathbb{E}_{s \sim d^\pi}\big[\lVert \nabla_a \hat{\bar{R}}_{\lambda, \theta}(s,\mu_\theta(s)) - \nabla_a \bar{R}(s, \mu_\theta(s))\lVert^2 \big]$:
$$\nabla_\theta J_\pi(\mu_\theta) = \mathbb{E}_{s \sim d^\pi}\big[\nabla_\theta \mu_\theta (s) \cdot \left. \nabla_a \bar{R}(s, a)\right|_{a = \mu_\theta(s)} \big] = \mathbb{E}_{s \sim d^\pi}\big[\nabla_\theta \mu_\theta (s) \cdot \left. \nabla_a \hat{\bar{R}}_{\lambda^*, \theta}(s, a)\right|_{a = \mu_\theta(s)} \big].$$
The use of compatible features requires each agent to observe not only the joint action taken  $a_{t+1} = (a_{t+1}^1, \dots, a_{t+1}^N)$ and the ``on-policy action'' $\dot{a}_{t+1} = (\dot{a}_{t+1}^1, \dots, \dot{a}_{t+1}^N)$, but also $\widetilde{a}_{t+1} = (\nabla_{\theta^1} \mu^1_{\theta^1_t}(s_{t+1}), \dots, \nabla_{\theta^N} \mu^N_{\theta^N_t}(s_{t+1}))$ (see the parts in blue in Algorithm~\ref{alg:Off_Policy}).

We illustrate  algorithm convergence on multi-agent extension of a continuous  bandit problem  from Sec. 5.1 of \cite{Silver:2014:DeterministicPolicyGradient}. Details  are in the Appendix.
Figure~\ref{fig} shows the convergence of Algorithms 1 and 2  averaged over 5 runs. In all cases,  the system converges and the agents are able to coordinate their actions to minimize system cost. 
\begin{figure}[!h]
\begin{subfigure}{.33\textwidth}
\includegraphics[width=\textwidth]{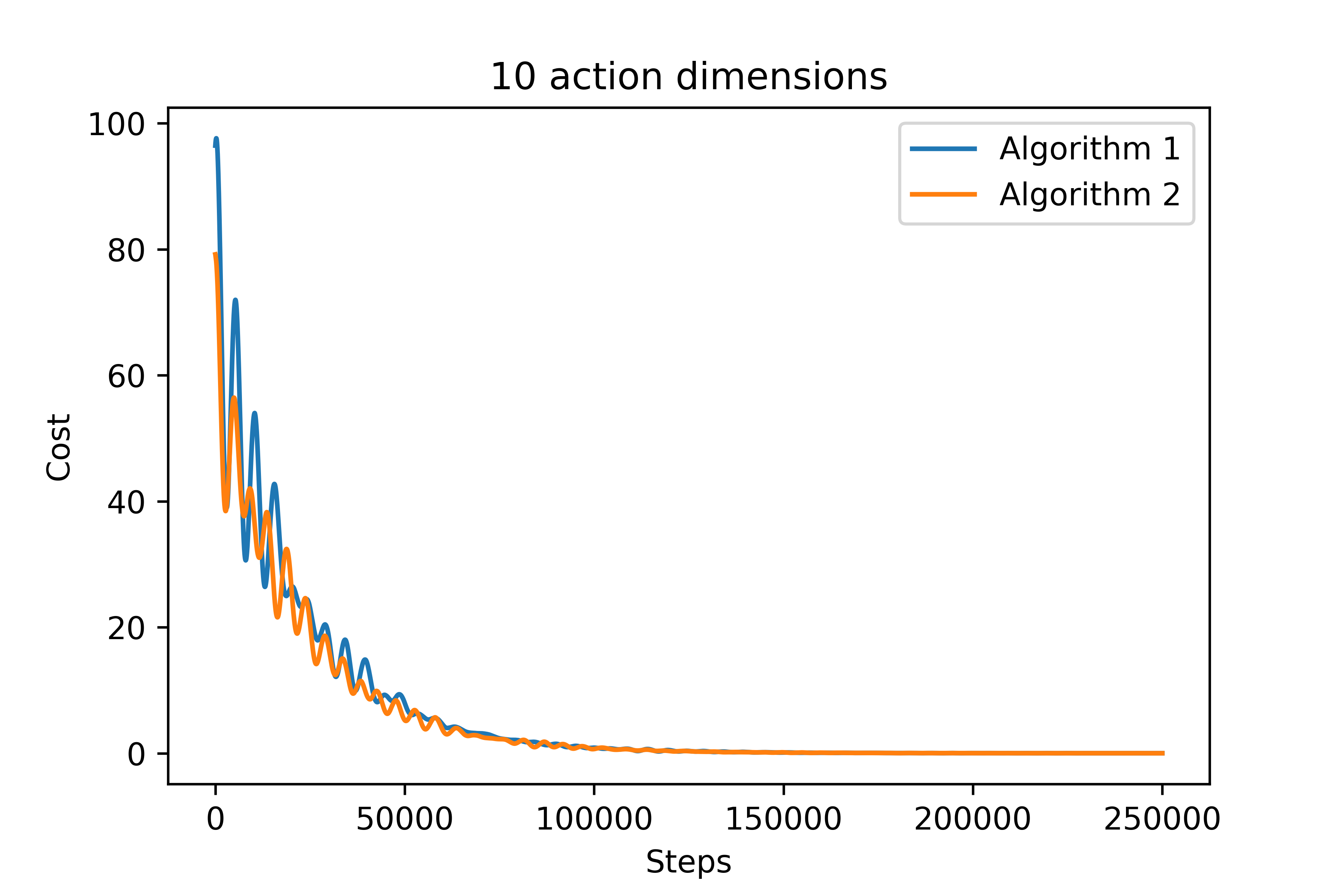}
\label{fig:10}
\end{subfigure}
\begin{subfigure}{.33\textwidth}
\includegraphics[width=\textwidth]{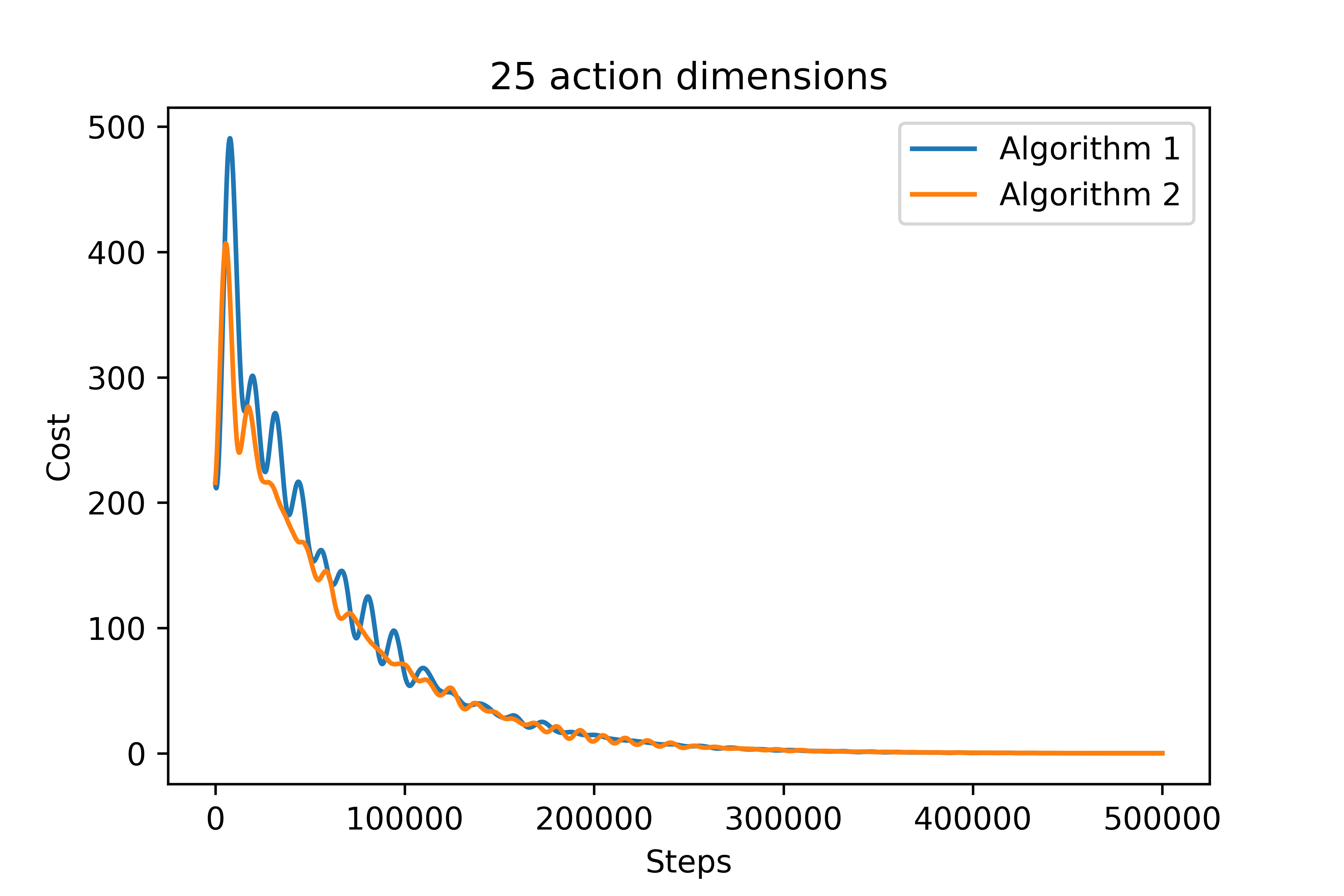}
\label{fig:25}
\end{subfigure}
\begin{subfigure}{.33\textwidth}
\includegraphics[width=\textwidth]{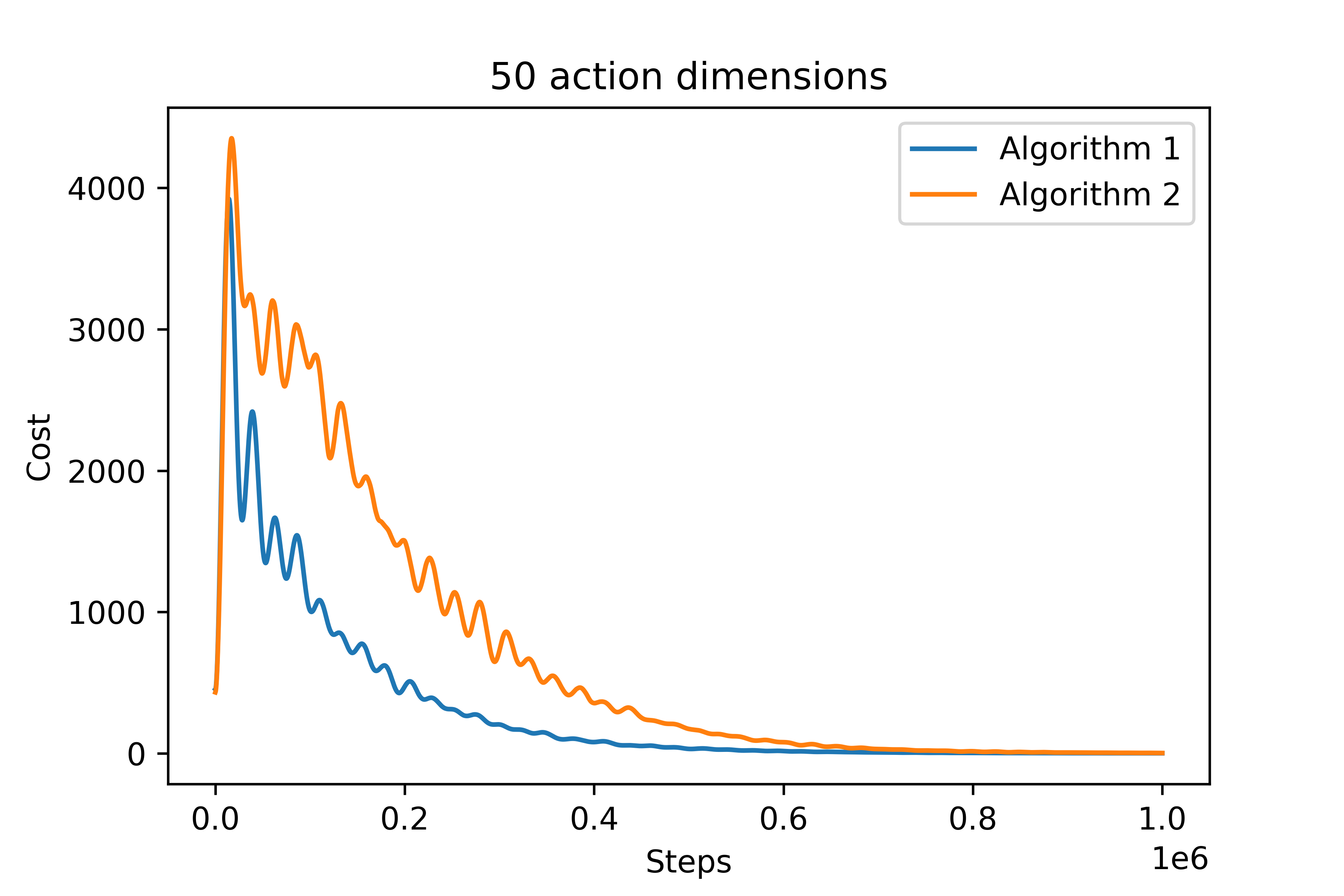}
\label{fig:50}
\end{subfigure}
\caption{Convergence of Algorithms 1 and 2 on the multi-agent continuous bandit problem.}
\label{fig}
\end{figure}

%% file: conclusion.tex
\section{Conclusion}

We have provided the tools needed to implement decentralized, deterministic actor-critic algorithms for  cooperative multi-agent reinforcement learning. We provide the expressions for the policy gradients, the algorithms themselves, and prove their convergence  in  on-policy and  off-policy settings. We also provide numerical results  for a continuous multi-agent bandit problem that demonstrates the convergence of our algorithms.
Our work differs from~\cite{DistributedOffPolicyActorCriticRL} as the latter was based on policy consensus whereas ours is based on critic consensus. Our approach represents  agreement between agents on every participants' contributions to the global reward, and as such,  provides a consensus scoring function with which to evaluate agents. Our approach may be used in compensation schemes to incentivize participation.
An interesting extension of this work would be to prove  convergence of our actor-critic algorithm for continuous state spaces, as 
it may hold with assumptions on the geometric ergodicity of the stationary state distribution induced by the deterministic policies (see   \cite{BorkarStochasticApprDynamicalSysViewpoint}). The  expected policy gradient (EPG) of  \cite{Ciosek:2018:ExpectedPG}, a hybrid  between  stochastic and deterministic policy gradient, would also be interesting to leverage.
The Multi-Agent Deep Deterministic Policy Gradient algorithm (MADDPG) of  \cite{lowe:2017:multi_AC_Mixed} assumes partial observability for each agent and would be a useful extension, but it is likely  difficult to extend our convergence guarantees to the partially observed setting.

%% file: supp.tex
\section{Appendix}

\subsection*{Numerical experiment details}

\begin{figure}[!h]
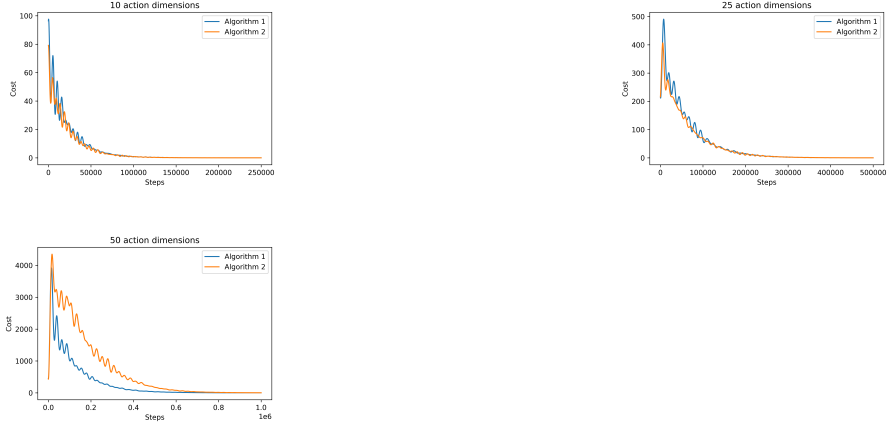

\begin{subfigure}{.33\textwidth}
\includegraphics[width=\textwidth]{m10_5runs}
\label{fig:10}
\end{subfigure}
\begin{subfigure}{.33\textwidth}
\includegraphics[width=\textwidth]{m25_5runs}
\label{fig:25}
\end{subfigure}
\begin{subfigure}{.33\textwidth}
\includegraphics[width=\textwidth]{m50_5runs}
\label{fig:50}
\end{subfigure}
\caption{Convergence of Algorithms 1 and 2 on the multi-agent continuous bandit problem.}
\label{fig}
\end{figure}

We demonstrate the convergence of our algorithm in a continuous bandit problem that is a multi-agent extension of the experiment in Section 5.1 of Silver et al. (2014). Each agent chooses an action $a^i\in\mathbb{R}^m$. We assume all agents have the same reward function given by $R^i(a) = -\left(\sum_i a^i - a^*\right)^\mathsf{T} C \left(\sum_i a^i - a^*\right)$. The matrix $C$ is positive definite with eigenvalues chosen from $\{0.1,1\}$, and $a^* = [4,\ldots,4]^\mathsf{T}$. We consider $10$ agents and action dimensions  $m = 10,20,50$. Note that there are multiple possible solutions for this problem, requiring the agents to coordinate their actions to sum to $a^*$.
We assume a target policy of the form $\mu_{\theta^i} = \theta^i$ for each agent $i$ and a Gaussian behaviour policy $\beta(\cdot)\sim\mathcal{N}(\theta^i,\sigma_\beta^2)$ where $\sigma_\beta = 0.1$. We use the Gaussian behaviour policy for both Algorithms 1 and 2. Strictly speaking, Algorithm 1 is on-policy, but in this simplified setting where the target policy is constant, the on-policy version would be degenerate such that the $Q$ estimate does not affect the TD-error. Therefore, we add a Gaussian behaviour policy to Algorithm 1.
Each agent maintains an estimate $Q^{\omega^i}(a)$ of the critic using a linear function of the compatible features $a - \theta$ and a bias feature. The critic is recomputed from each successive batch of $2m$ steps and the actor is updated once per batch. The critic step size is $0.1$ and the actor step size is $0.01$. Performance is evaluated by measuring the cost of the target policy (without exploration). 
Figure~\ref{fig} shows the convergence of Algorithms 1 and 2  averaged over 5 runs. In all cases,  the system converges and the agents are able to coordinate their actions to minimize system cost. The jupyter notebook will be made available for others to use. In fact, in this simple experiment, we also observe convergence under discounted rewards.

\subsection*{Assumptions}

\begin{asm}[Linear approximation, average-reward] \label{asm_compatible_features_R}
For each agent $i$, the average-reward function $\bar{R}$ is parameterized by the class of linear functions, i.e., $\hat{\bar{R}}_{\lambda^i, \theta}(s,a) = w_\theta(s,a) \cdot \lambda^i$ where $w_\theta(s,a) = \big[w_{\theta,1}(s,a), \dots, w_{\theta,K}(s,a) \big] \in  \mathbb{R}^K$ is the feature associated with the state-action pair $(s, a)$. The feature vectors $w_\theta(s,a)$, as well as $\nabla_a w_{\theta, k} (s,a)$ are uniformly bounded for any $s \in \mathcal{S}$, $a \in \mathcal{A}, k \in \pabb{1, K}$. Furthermore, we assume that the feature matrix $W_\pi \in \mathbb{R}^{|S| \times K}$ has full column rank, where the $k$-th column of $W_{\pi, \theta}$ is $\big[\int_\mathcal{A}\pi(a|s)w_{\theta, k}(s, a)\textnormal{d}a, s \in \mc{S}\big]$ for any $k \in \llbracket1,K\rrbracket$.
\end{asm}

\begin{asm}[Linear approximation, action-value] \label{Linear_appr}
For each agent $i$, the action-value function is parameterized by the class of linear functions, i.e., $\hat{Q}_{\omega^i}(s,a) = \phi(s,a) \cdot \omega^i$ where $\phi(s,a) = \big[\phi_1(s,a), \dots, {\phi_K(s,a) \big]} \in  \mathbb{R}^K$ is the feature associated with the state-action pair $(s, a)$. The feature vectors $\phi(s,a)$, as well as $\nabla_a \phi_k (s,a)$ are uniformly bounded for any $s \in \mathcal{S}$, $a \in \mathcal{A}, k \in \{1, \dots, K\}$. Furthermore, we assume that for any $\theta \in \Theta$, the feature matrix $\Phi_\theta \in \mathbb{R}^{\vert \mathcal{S} \vert \times K}$ has full column rank, where the $k$-th column of $\Phi_\theta$ is $\big[\phi_k(s, \mu_\theta(s)), s \in \mathcal{S}\big]$ for any $k \in \llbracket1, K\rrbracket$. Also, for any $u \in \mathbb{R}^K$, $\Phi_\theta u \neq \textbf{1}$.
\end{asm}

\begin{asm}[Bounding $\theta$] \label{Bound_theta}
The update of the policy parameter $\theta^i$ includes a local projection by $\Gamma^i:\mathbb{R}^{m_i} \to \Theta^i$ that projects any $\theta^i_t$ onto a compact set $\Theta^i$ that can be expressed as $\{\theta^i| q_j^i(\theta^i) \leq 0, \ j = 1, \dots, s^i\} \subset \mathbb{R}^{m_i}$, for some real-valued, continuously differentiable functions $\{q^i_j\}_{1 \leq j \leq s^i}$ defined on $\mathbb{R}^{m_i}$. We also assume that $\Theta = \prod_{i=1}^N \Theta^i$ is large enough to include at least one local minimum of $J(\theta)$.
\end{asm}

We use $\{\mathcal{F}_t\}$ to denote the filtration with $\mathcal{F}_t = \sigma(s_\tau, C_{\tau-1}, a_{\tau-1}, r_{\tau-1}, \tau \leq t)$.

\begin{asm}[Random matrices] \label{asm_random_matrix}
The sequence of non-negative random matrices $\{C_t = (c_t^{ij})_{ij}\}$ satisfies:
\begin{enumerate}
    \item $C_t$ is row stochastic and $\mathbb{E}(C_t|\mathcal{F}_t)$ is a.s. column stochastic for each $t$, i.e., $C_t \textbf{1} = \textbf{1}$ and $\textbf{1}^\top\mathbb{E}(C_t|\mathcal{F}_t) = \textbf{1}^\top$ a.s. Furthermore, there exists a constant $\eta \in (0, 1)$ such that, for any $c_t^{ij} > 0$, we have $c_t^{ij} \geq \eta$.
    \item $C_t$ respects the communication graph $\mathcal{G}_t$, i.e., $c_t^{ij} = 0$ if $(i, j) \notin \mathcal{E}_t$.
    \item The spectral norm of $\mathbb{E}\big[C_t^\top \cdot (I - \textbf{1} \textbf{1}^\top / N) \cdot C_t \big]$ is smaller than one.
    \item Given the $\sigma$-algebra generated by the random variables before time $t$, $C_t$, is conditionally independent of $s_t, a_t$ and $r^i_{t+1}$ for any $i \in \mathcal{N}$.
\end{enumerate}
\end{asm}

\begin{asm} [Step size rules, on-policy] \label{Two_timesteps}
The stepsizes $\beta_{\omega, t}, \beta_{\theta, t}$ satisfy:
\begin{align*}
    &\sum_t \beta_{\omega, t} = \sum_t \beta_{\theta, t} = \infty \\
    &\sum_t (\beta_{\omega, t}^2 + \beta_{\theta, t}^2) < \infty\\
    &\sum_t |\beta_{\theta, t+1} - \beta_{\theta, t}| < \infty.
\end{align*}
In addition, $\beta_{\theta, t} = o(\beta_{\omega, t})$ and $\text{lim}_{t \to \infty} \beta_{\omega, t+1}/\beta_{\omega, t} = 1$.
\end{asm}

\begin{asm}[Step size rules, off-policy] \label{asm_algo2_two_timesteps}
The step-sizes $\beta_{\lambda, t}$, $\beta_{\theta, t}$ satisfy:
\begin{alignat*}{2}
&\sum_t \beta_{\lambda, t} = \sum_t \beta_{\theta,t} = \infty, \hspace{4em} &\sum_t \beta_{\lambda, t}^2 + \beta_{\theta, t}^2 < \infty\\
&\beta_{\theta, t} = o(\beta_{\lambda, t}), &\underset{t\to \infty}{\textnormal{lim}}\beta_{\lambda, t+1} / \beta_{\lambda, t} = 1.
\end{alignat*}
\end{asm}


\subsection*{Proof of Theorem \ref{thm:On_Policy_Grad}} \label{app:On_Policy_Grad}

The proof follows the same scheme as \cite{Policy_grad_meth_RL_fct_appr}, naturally extending their results for a deterministic policy $\mu_\theta$ and a continuous action space $\mc{A}$. 

Note that our regularity assumptions 
ensure that, for any $s \in \mathcal{S}$, $V_\theta(s)$, $\nabla_\theta V_\theta(s)$, $J(\theta)$, $\nabla_\theta J(\theta)$, $d^\theta$(s) are Lipschitz-continuous functions of $\theta$ (since $\mu_\theta$ is twice continuously differentiable and $\Theta$ is compact), and that $Q_\theta(s,a)$ and $\nabla_a Q_\theta(s,a)$ are Lipschitz-continuous functions of $a$ (\cite{SimuBasedOptMDP}).

We first show that $\nabla_\theta J(\theta) = \mathbb{E}_{s \sim d^\theta} \big[\nabla_\theta \mu_\theta(s) \nabla_a \left. Q_\theta(s,a)\right|_{a = \mu_\theta(s)}]$.

The Poisson equation under policy $\mu_\theta$ is given by \cite{Puterman:1994:MDP}
\begin{equation*}
    Q_\theta(s,a) = \bar{R}(s, a) - J(\theta) + \sum_{s^\prime \in \mathcal{S}}  P(s^\prime \vert s, a) V_\theta(s^\prime) \label{Q_R_J_V}.
\end{equation*}
So,
\begin{align*}
    \nabla_\theta V_\theta(s) &= \nabla_\theta Q_\theta(s, \mu_\theta(s)) \\
    &= \nabla_\theta \big[\bar{R}(s, \mu_\theta(s)) - J(\theta) + \sum_{s^\prime \in \mathcal{S}} P(s^\prime \vert s, \mu_\theta(s)) V_\theta(s^\prime) \big] \\
    &= \nabla_\theta \mu_\theta (s) \left. \nabla_a \bar{R}(s,a) \right|_{a = \mu_\theta(s)} - \nabla_\theta J(\theta) + \nabla_\theta \sum_{s^\prime \in \mathcal{S}} P(s^\prime \vert s, \mu_\theta(s)) V_\theta(s^\prime) \\
        &= \nabla_\theta \mu_\theta (s) \left. \nabla_a \bar{R}(s,a) \right|_{a = \mu_\theta(s)} - \nabla_\theta J(\theta)
        \\
        &\quad +  \sum_{s^\prime \in \mathcal{S}} \nabla_\theta \mu_\theta (s) \left. \nabla_a P(s^\prime \vert s, a) \right|_{a = \mu_\theta(s)} V_\theta (s^\prime) 
        + \sum_{s^\prime \in \mathcal{S}} P(s^\prime \vert s, \mu_\theta(s)) \nabla_\theta V_\theta(s^\prime)
    \\
    &= \nabla_\theta \mu_\theta (s) \nabla_a \left.\Big[\bar{R}(s, a) + \sum_{s^\prime \in \mathcal{S}}  P(s \vert s^\prime,a) V_\theta (s^\prime) \Big]\right|_{a = \mu_\theta(s)} 
    \\
    &\quad - \nabla_\theta J(\theta) + \sum_{s^\prime \in \mathcal{S}} P(s^\prime \vert s, \mu_\theta(s)) \nabla_\theta V_\theta(s^\prime)\\
    &= \nabla_\theta \mu_\theta(s) \nabla_a \left. Q_\theta(s,a)\right|_{a = \mu_\theta(s)}  + \sum_{s^\prime \in \mathcal{S}} P(s^\prime \vert s, \mu_\theta(s)) \nabla_\theta V_\theta(s^\prime) - \nabla_\theta J(\theta)
\end{align*}
Hence,
\begin{align*}
    \nabla_\theta J(\theta) &= \nabla_\theta \mu_\theta(s) \nabla_a \left. Q_\theta(s,a)\right|_{a = \mu_\theta(s)}  + \sum_{s^\prime \in \mathcal{S}} P(s^\prime \vert s, \mu_\theta(s)) \nabla_\theta V_\theta(s^\prime) - \nabla_\theta V_\theta(s) 
    \\
        \sum_{s \in \mathcal{S}} d^\theta(s) \nabla_\theta J(\theta) &= \sum_{s \in \mathcal{S}} d^\theta(s) \nabla_\theta \mu_\theta(s) \nabla_a \left. Q_\theta(s,a)\right|_{a = \mu_\theta(s)}
        \\
        &\quad + \sum_{s \in \mathcal{S}} d^\theta(s) \sum_{s^\prime \in \mathcal{S}} P(s^\prime \vert s, \mu_\theta(s)) \nabla_\theta V_\theta(s^\prime) - \sum_{s \in \mathcal{S}} d^\theta(s) \nabla_\theta V_\theta(s).
\end{align*}
Using stationarity property of $d^\theta$, we get 
$$\sum_{s \in \mathcal{S}} \sum_{s^\prime \in \mathcal{S}} d^\theta (s) P(s^\prime \vert s, \mu_\theta(s)) \nabla_\theta V_\theta(s^\prime) = \sum_{ s^\prime \in \mathcal{S}} d^\theta(s^\prime) \nabla_\theta V_\theta(s^\prime).$$
Therefore, we get
$$\nabla_\theta J(\theta) = \sum_{s \in \mathcal{S}} d^\theta(s) \nabla_\theta \mu_\theta(s) \left. \nabla_a Q_\theta(s,a)\right|_{a = \mu_\theta(s)} = \mathbb{E}_{s \sim d^\theta} \big[\nabla_\theta \mu_\theta(s) \left. \nabla_a Q_\theta(s,a)\right|_{a = \mu_\theta(s)}].$$

Given that $\nabla_{\theta^i} \mu_{\theta}^j (s) = 0$ if $i \neq j$, we have $\nabla_\theta \mu_\theta(s) = \text{Diag}(\nabla_{\theta^1} \mu_{\theta_1}^1 (s), \dots, \nabla_{\theta^N} \mu_{\theta_N}^N (s))$, which implies
\begin{equation} \label{local_grad_J_Q}
    \nabla_{\theta^i}J(\theta) = \mathbb{E}_{s \sim d^\theta} \big[\nabla_{\theta^i} \mu^i_{\theta^i}(s) \nabla_{a^i} \left. Q_\theta(s, \mu^{-i}_{\theta^{-i}}(s), a^i)\right|_{a^i = \mu^i_{\theta^i}(s)}].
\end{equation}


\subsection*{Proof of Theorem \ref{thm:limit_stoch_grad}} 
\label{app:limit_stoch_grad}

We extend the notation for off-policy reward function to stochastic policies as follows. Let $\beta$ be a behavior policy under which $\{s_t\}_{t\geq 0}$ is irreducible and aperiodic, with stationary distribution $d^\beta$. For a stochastic policy $\pi:\mc{S}\to\mc{P}(\mc{A})$, we define
\begin{equation*}
    J_\beta(\pi) = \sum_{s \in \mathcal{S}} d^\beta(s) \int_{\mathcal{A}} \pi(a|s) \bar{R}(s, a) \textnormal{d}a.
\end{equation*}
Recall that for a deterministic policy $\mu:\mc{S}\to\mc{A}$, we have
\begin{equation*}
    J_\beta(\mu) = \sum_{s \in \mathcal{S}} d^\beta(s) \bar{R}(s, \mu(s)).
\end{equation*}




We introduce the following conditions which are identical to \textbf{Conditions B1} from \cite{DeterministicPolicyGradient}.

\begin{cond} \label{cond:regular_delta_appr}
Functions $\nu_\sigma$ parametrized by $\sigma$ are said to be regular delta-approximation on $\mathcal{R} \subset \mathcal{A}$ if they satisfy the following conditions:
\begin{enumerate}
    \item The distributions $\nu_\sigma$ converge to a delta distribution: $\textnormal{lim}_{\sigma \downarrow 0} \int_\mathcal{A} \nu_\sigma(a^\prime, a) f(a) \textnormal{d}a = f(a^\prime)$ for $a^\prime \in \mathcal{R}$ and suitably smooth $f$. Specifically we require that this convergence is uniform in $a^\prime$ and over any class $\mathcal{F}$ of $L$-Lipschitz and bounded functions, $\lVert \nabla_a f(a)\lVert < L < \infty$, $\textnormal{sup}_a f(a) < b < \infty$, i.e.:
    \begin{equation*}
        \underset{\sigma \downarrow 0}{\textnormal{lim }} \underset{f \in \mathcal{F}, a^\prime \in \mc{R}}{\textnormal{sup }} \left|\int_\mathcal{A} \nu_\sigma(a^\prime, a) f(a) \textnormal{d}a - f(a^\prime) \right| = 0.
    \end{equation*}
    \item For each $a^\prime \in \mathcal{R}$, $\nu_\sigma(a^\prime, \cdot)$ is supported on some compact $\mathcal{C}_{a^\prime} \subseteq \mathcal{A}$ with Lipschitz boundary $\textnormal{bd} (\mathcal{C}_{a^\prime})$, vanishes on the boundary and is continuously differentiable on $\mathcal{C}_{a^\prime}$.
    \item For each $a^\prime \in \mathcal{R}$, for each $a \in \mathcal{A}$, the gradient $\nabla_{a^\prime} \nu_\sigma(a^\prime, a)$ exists.
    \item Translation invariance: for all $a \in \mathcal{A}, a^\prime \in \mathcal{R}$, and any $\delta \in \mathbb{R}^n$ such that $a + \delta \in \mathcal{A}$, $a^\prime + \delta \in \mathcal{A}$, $\nu_\sigma(a^\prime, a) = \nu_\sigma(a^\prime + \delta, a + \delta)$.
\end{enumerate}
\end{cond}

The following lemma is an immediate corollary of \textbf{Lemma 1} from \cite{DeterministicPolicyGradient}.
\begin{lem}\label{grad_nu}
Let $\nu_\sigma$ be a regular delta-approximation on $\mathcal{R} \subseteq \mathcal{A}$. Then, wherever the gradients exist
$$\nabla_{a^\prime}\nu(a^\prime, a) = - \nabla_a \nu(a^\prime, a).$$
\end{lem}

Theorem \ref{thm:limit_stoch_grad} is a less technical restatement of the following result.

\begin{thm}
\label{thm:limit_stoch_grad_technical}
Let $\mu_\theta: \mathcal{S} \to \mathcal{A}$. Denote the range of $\mu_\theta$ by $\mathcal{R}_\theta \subseteq \mathcal{A}$, and $\mathcal{R} = \cup_\theta \mathcal{R}_\theta$. For each $\theta$, consider $\pi_{\theta, \sigma}$ a stochastic policy such that $\pi_{\theta, \sigma}(a|s) = \nu_\sigma(\mu_\theta(s),a)$, where $\nu_\sigma$ satisfy Conditions \ref{cond:regular_delta_appr} on $\mathcal{R}$. 
Then, there exists $r > 0$ such that, for each $\theta \in \Theta$, $\sigma \mapsto J_{\pi_{\theta, \sigma}}(\pi_{\theta, \sigma})$, $\sigma \mapsto J_{\pi_{\theta, \sigma}}(\mu_\theta)$, $\sigma \mapsto \nabla_\theta J_{\pi_{\theta, \sigma}}(\pi_{\theta, \sigma})$, and $\sigma \mapsto \nabla_\theta J_{\pi_{\theta, \sigma}}(\mu_\theta)$ are properly defined on $\big[0, r\big]$ (with $J_{\pi_{\theta, 0}}(\pi_{\theta, 0}) = J_{\pi_{\theta, 0}}(\mu_\theta) = J_{\mu_\theta}(\mu_\theta)$ and $\nabla_\theta J_{\pi_{\theta, 0}}(\pi_{\theta, 0}) = \nabla_\theta J_{\pi_{\theta, 0}}(\mu_\theta) = \nabla_\theta J_{\mu_\theta}(\mu_\theta)$), and we have:
\begin{equation*}
    \underset{\sigma \downarrow 0}{\textnormal{lim }} \nabla_\theta J_{\pi_{\theta, \sigma}} (\pi_{\theta, \sigma}) = \underset{\sigma \downarrow 0}{\textnormal{lim }} \nabla_\theta J_{\pi_{\theta, \sigma}}(\mu_\theta) = \nabla_\theta J_{\mu_\theta}(\mu_\theta).
\end{equation*}
\end{thm}

To prove this result, we first state and prove the following Lemma.


\begin{lem} \label{stationary_distrib}
There exists $r > 0$ such that, for all $\theta \in \Theta$ and $\sigma \in \big[0, r\big]$, stationary distribution $d^{\pi_{\theta, \sigma}}$ exists and is unique. Moreover, for each $\theta \in \Theta$, $\sigma \mapsto d^{\pi_{\theta, \sigma}}$ and $\sigma \mapsto \nabla_\theta d^{\pi_{\theta, \sigma}}$ are properly defined on $\big[0, r\big]$ and both are continuous at $0$.
\end{lem}

\begin{proof}[Proof of Lemma \ref{stationary_distrib}]
For any policy $\beta$, we let $\left(P^\beta_{s,s^\prime}\right)_{s, s^\prime \in \mathcal{S}}$ be the transition matrix associated to the Markov Chain $\{s_t\}_{t \geq 0}$ induced by $\beta$. In particular, for each $\theta \in \Theta$, $\sigma >0$, $s, s^\prime \in \mathcal{S}$, we have
\begin{align*}
    P^{\mu_\theta}_{s, s^\prime} 
    &= P(s^\prime |s, \mu_\theta(s)),
    \\
    P^{\pi_{\theta, \sigma}}_{s, s^\prime} 
    &= \int_\mathcal{A} \pi_{\theta, \sigma} (a | s) P(s^\prime |s, a) \textnormal{d}a = \int_\mathcal{A} \nu_\sigma(\mu_\theta(s), a) P(s^\prime | s, a) \textnormal{d}a.
\end{align*}
Let $\theta \in \Theta$, $s, s^\prime \in \mathcal{S}$,  $\left(\theta_n\right)\in\Theta^\mathbb{N}$ such that $\theta_n \rightarrow \theta$ and $\left(\sigma_n\right)_{n \in \mathbb{N}} \in {\mathbb{R}^+}^\mathbb{N}$, $\sigma_n \downarrow 0$:
\begin{equation*}
    \left|P^{\pi_{\theta_n, \sigma_n}}_{s, s^\prime} - P^{\mu_\theta}_{s, s^\prime}\right| \leq \left|P^{\pi_{\theta_n, \sigma_n}}_{s, s^\prime} - P^{\mu_{\theta_n}}_{s, s^\prime}\right| + \left|P^{\mu_{\theta_n}}_{s, s^\prime} - P^{\mu_\theta}_{s, s^\prime}\right|.\\
\end{equation*}
Applying the first condition of Conditions \ref{cond:regular_delta_appr} with $f: a \mapsto P(s^\prime | s, a)$ belonging to $\mathcal{F}$: 
\begin{align*}
    \left|P^{\pi_{\theta_n, \sigma_n}}_{s, s^\prime} - P^{\mu_{\theta_n}}_{s, s^\prime}\right| &= \left|\int_\mathcal{A} \nu_{\sigma_n}(\mu_{\theta_n}(s), a) P(s^\prime | s, a) \textnormal{d}a - P(s^\prime | s, \mu_{\theta_n}(s)) \right|\\
    &\leq \underset{f \in \mathcal{F}, a^\prime \in \mathcal{R}}{\sup} \left|\int_\mathcal{A} \nu_{\sigma_n}(a^\prime, a) f(a) \textnormal{d}a - f(a^\prime) \right|
    \underset{n \rightarrow \infty}{\longrightarrow} 0.
\end{align*}
By regularity assumptions on $\theta \mapsto \mu_\theta(s)$ 
and $P(s^\prime|s, \cdot)$, 
we have
\begin{equation*}
    \left|P^{\mu_{\theta_n}}_{s, s^\prime} - P^{\mu_\theta}_{s, s^\prime}\right| = \left| P(s^\prime | s, \mu_{\theta_n}(s)) - P(s^\prime | s, \mu_\theta(s))\right| \underset{n \rightarrow \infty}{\longrightarrow} 0.
\end{equation*}
Hence,
\begin{equation*}
    \left|P^{\pi_{\theta_n, \sigma_n}}_{s, s^\prime} - P^{\mu_\theta}_{s, s^\prime}\right| \underset{n \rightarrow \infty}{\longrightarrow} 0.
\end{equation*}
Therefore, for each $s, s^\prime \in \mathcal{S}$, $(\theta, \sigma) \mapsto P^{\pi_{\theta, \sigma}}_{s, s^\prime}$, with $P^{\pi_{\theta, 0}}_{s, s^\prime} = P^{\mu_\theta}_{s, s^\prime}$, is continuous on $\Theta \times \{0\}$. Note that, for each $n \in \mathbb{N}$, $P \mapsto \prod_{s, s^\prime} \left(P^n\right)_{s, s^\prime}$ is a polynomial function of the entries of $P$. Thus, for each $n \in \mathbb{N}$, $f_n: (\theta, \sigma) \mapsto \prod_{s, s^\prime} \left({P^{\pi_{\theta, \sigma}}}^n\right)_{s, s^\prime}$, with $f_n(\theta, 0) = \prod_{s, s^\prime} \left({P^{\mu_\theta}}^n\right)_{s, s^\prime}$ is continuous on $\Theta \times \{0\}$. Moreover, for each $\theta \in \Theta, \sigma \geq 0$, from the structure of $P^{\pi_{\theta, \sigma}}$, if there is some $n^* \in \mathbb{N}$ such that $f_{n^*}(\theta, \sigma) > 0$ then, for all $n \geq n^*$, $f_n(\theta, \sigma) > 0$.

Now let us suppose that there exists $\left(\theta_n\right)\in\Theta^\mathbb{N^*}$ such that, for each $n > 0$ there is a $\sigma_n \leq n^{-1}$ such that $f_n(\theta_n, \sigma_n) = 0$. By compacity of $\Theta$, we can take $\left(\theta_n\right)$ converging to some $\theta \in \Theta$. For each $n^* \in \mathbb{N}$, by continuity we have $f_{n^*}(\theta, 0) = \underset{n \to \infty}{\lim} f_{n^*}(\theta_n, \sigma_n) = 0$. 
Since $P^{\mu_\theta}$ is irreducible and aperiodic, there is some $n \in \mathbb{N}$ such that for all $s, s^\prime \in \mathcal{S}$ and for all $n^* \geq n$, $\left({P^{\mu_\theta}}^{n^*}\right)_{s, s^\prime} > 0$, i.e. $f_{n^*}(\theta, 0) > 0$. This leads to a contradiction.

Hence, there exists $n^* > 0$ such that for all $\theta \in \Theta$ and $\sigma \leq {n^*}^{-1}$, $f_n(\theta, \sigma) > 0$. We let $r = {n^*}^{-1}$. It follows that, for all $\theta \in \Theta$ and $\sigma \in \big[0, r\big]$, $P^{\pi_{\theta, \sigma}}$ is a transition matrix associated to an irreducible and aperiodic Markov Chain, thus $d^{\pi_{\theta, \sigma}}$ is well defined as the unique stationary probability distribution associated to $P^{\pi_{\theta, \sigma}}$. We fix $\theta \in \Theta$ in the remaining of the proof.
\bigbreak
Let $\beta$ a policy for which the Markov Chain corresponding to $P^\beta$ is irreducible and aperiodic. Let $s_* \in \mathcal{S}$, as asserted in \cite{SimuBasedOptMDP}, considering stationary distribution $d^\beta$ as a vector $\left(d^{\beta}_s\right)_{s \in \mathcal{S}} \in \mathbb{R}^{|\mathcal{S}|}$, $d^\beta$ is the unique solution of the balance equations:
\begin{align*}
    \sum_{s \in \mathcal{S}} d^{\beta}_s P^{\beta}_{s, s^\prime} &= d^\beta_{s^\prime} \quad s^\prime \in \mathcal{S}\backslash\{s_*\},\\
    \sum_{s \in \mathcal{S}} d^{\beta}_s &= 1.
\end{align*}

Hence, we have $A^\beta$ an $|\mathcal{S}| \times |\mathcal{S}|$ matrix and $a \neq 0$ a constant vector of $\mathbb{R}^{|\mathcal{S}|}$ such that the balance equations is of the form
\begin{equation} \label{balance_equation}
    A^\beta d^{\beta} = a
\end{equation}
with $A^\beta_{s, s^\prime}$ depending on $P^{\beta}_{s^\prime, s}$ in an affine way, for each $s, s^\prime \in \mathcal{S}$. Moreover, $A^\beta$ is invertible, thus $d^\beta$ is given by
\begin{equation*}
    d^\beta = \frac{1}{\det(A^\beta)} \textnormal{adj}(A^\beta)^\top a.
\end{equation*}
Entries of $\textnormal{adj}(A^\beta)$ and $\det(A^\beta)$ are polynomial functions of the entries of $P^\beta$.
\bigbreak

Thus, $\sigma \mapsto d^{\pi_{\theta, \sigma}} = \frac{1}{\det(A^{\pi_{\theta, \sigma}})} \textnormal{adj}(A^{\pi_{\theta, \sigma}})^\top a$ is defined on $\big[0, r\big]$ and is continuous at 0.
\bigbreak

Lemma \ref{grad_nu} and integration by parts imply that, for $s, s^\prime \in \mathcal{S}$, $\sigma \in \big[0, r\big]$:
\begin{align*}
\int_\mathcal{A} \left. \nabla_{a^\prime} \nu_\sigma(a^\prime, a) \right|_{a^\prime = \mu_\theta(s)} P(s^\prime | s, a) \textnormal{d}a &= - \int_\mathcal{A} \nabla_{a} \nu_\sigma(\mu_\theta(s), a) P(s^\prime | s, a) \textnormal{d}a \\
&= \int_{\mathcal{C}_{\mu_\theta(s)}} \nu_\sigma(\mu_\theta(s), a) \nabla_a P(s^\prime |s, a) \textnormal{d}a + \textnormal{boundary terms} \\
&=  \int_{\mathcal{C}_{\mu_\theta(s)}} \nu_\sigma(\mu_\theta(s), a) \nabla_a P(s^\prime |s, a) \textnormal{d}a
\end{align*}
where the boundary terms are zero since $\nu_\sigma$ vanishes on the boundary due to Conditions \ref{cond:regular_delta_appr}.

Thus, for $s, s^\prime \in \mathcal{S}$, $\sigma \in \big[0, r\big]$:
\begin{align}
\nabla_\theta P^{\pi_{\theta, \sigma}}_{s, s^\prime} &= \nabla_\theta \int_\mathcal{A} \pi_{\theta, \sigma} (a | s) P(s^\prime |s, a) \textnormal{d}a \nonumber\\
&= \int_\mathcal{A} \nabla_\theta \pi_{\theta, \sigma}(a |s) P(s^\prime | s, a) \textnormal{d}a \label{exchange_grad_int}\\
&= \int_\mathcal{A} \nabla_\theta \mu_\theta(s) \left. \nabla_{a^\prime} \nu_\sigma(a^\prime, a) \right|_{a^\prime = \mu_\theta(s)} P(s^\prime | s, a) \textnormal{d}a \nonumber \\
&= \nabla_\theta \mu_\theta(s) \int_{\mathcal{C}_{\mu_\theta(s)}} \nu_\sigma(\mu_\theta(s), a) \nabla_a P(s^\prime |s, a) \textnormal{d}a \nonumber
\end{align}

where exchange of derivation and integral in (\ref{exchange_grad_int}) follows by application of Leibniz rule with:
\begin{itemize}
    \item $\forall a \in \mathcal{A}$, $\theta \mapsto \pi_{\theta, \sigma} (a |s) P(s^\prime | s, a)$ is differentiable, and $\nabla_\theta \pi_{\theta, \sigma} (a |s) P(s^\prime | s, a) = \nabla_\theta \mu_\theta(s) \left. \nabla_{a^\prime} \nu_\sigma(a^\prime, a)\right|_{a^\prime = \mu_\theta(s)}$.\\
    \item Let $a^* \in \mathcal{R}$, $\forall \theta \in \Theta$, 
    \begin{align}
        \left\lVert \nabla_\theta \pi_{\theta, \sigma} (a |s) P(s^\prime | s, a) \right\lVert &= \left\lVert \nabla_\theta \mu_\theta(s) \left. \nabla_{a^\prime} \nu_\sigma(a^\prime, a)\right|_{a^\prime = \mu_\theta(s)} \right\lVert\nonumber\\
        &\leq \left\lVert\nabla_\theta\mu_\theta(s) \right\lVert_\textnormal{op} \left\lVert\left. \nabla_{a^\prime} \nu_\sigma(a^\prime, a)\right|_{a^\prime = \mu_\theta(s)}\right\lVert\nonumber \nonumber \\
        &\leq \underset{\theta \in \Theta}{\sup} \left\lVert\nabla_\theta\mu_\theta(s) \right\lVert_\textnormal{op} \left\lVert \nabla_a \nu_\sigma(\mu_\theta(s), a)\right\lVert\nonumber\\
        &= \underset{\theta \in \Theta}{\sup} \left\lVert\nabla_\theta\mu_\theta(s) \right\lVert_\textnormal{op} \left\lVert \nabla_a \nu_\sigma(a^*, a - \mu_\theta(s) + a^*)\right\lVert \label{align_delta}\\
        &\leq \underset{\theta \in \Theta}{\sup} \left\lVert\nabla_\theta\mu_\theta(s) \right\lVert_\textnormal{op} \underset{a \in \mathcal{C}_{a^*}}{\sup} \left\lVert\nabla_a \nu_\sigma(a^*, a)\right\lVert \textbf{1}_{a \in \mathcal{C}_{a^*}} \nonumber
    \end{align}
    where $\lVert \cdot \lVert_\textnormal{op}$ denotes the operator norm, and (\ref{align_delta}) comes from translation invariance (we take $\nabla_a \nu_\sigma(a^*, a) = 0$ for $a \in \mathbb{R}^n\backslash \mathcal{C}_{a^*}$). $a \mapsto \underset{\theta \in \Theta}{\sup} \left\lVert\nabla_\theta\mu_\theta(s) \right\lVert_\textnormal{op} \underset{a \in \mathcal{C}_{a^*}}{\sup} \left\lVert\nabla_a \nu_\sigma(a^*, a)\right\lVert \textbf{1}_{a \in \mathcal{C}_{a^*}}$ is measurable, bounded and supported on $\mathcal{C}_{a^*}$, so it is integrable on $\mathcal{A}$.
    \item Dominated convergence ensures that, for each $k \in \llbracket 1, m\rrbracket$, partial derivative $g_k(\theta) = \partial_{\theta_k} \int_\mathcal{A} \nabla_\theta \pi_{\theta, \sigma} (a |s) P(s^\prime | s, a) \textnormal{d}a$ is continuous: let $\theta_n \downarrow \theta$, then
    \begin{align*}
        g_k(\theta_n) &= \partial_{\theta_k} \int_\mathcal{A} \nabla_\theta \pi_{\theta_n, \sigma} (a |s) P(s^\prime | s, a) \textnormal{d}a\\
        &= \partial_{\theta_k} \mu_{\theta_n} (s) \int_{\mathcal{C}_{a^*}}  \nu_\sigma(a^*, a - \mu_{\theta_n}(s) + a^*) \nabla_a P(s^\prime | s,a) \textnormal{d}a\\
        &\underset{n \to \infty}{\longrightarrow}
        \partial_{\theta_k} \mu_{\theta} (s)  \int_{\mathcal{C}_{a^*}}  \nu_\sigma(a^*, a - \mu_{\theta}(s) + a^*) \nabla_a P(s^\prime | s,a) \textnormal{d}a = g_k(\theta)
    \end{align*}
    with the dominating function $a \mapsto \underset{a \in \mathcal{C}_{a^*}}{\sup} |\nu_\sigma(a^*,a)| \underset{a \in \mathcal{A}}{\sup}\left\lVert\nabla_a P(s^\prime|s,a) \right\lVert \textbf{1}_{a \in \mathcal{C}_{a^*}}$.
    \end{itemize}
\bigbreak

Thus $\sigma \mapsto \nabla_\theta P^{\pi_{\theta, \sigma}}_{s, s^\prime}$ is defined for $\sigma \in \big[0, r\big]$ and is continuous at 0, with $\nabla_\theta P^{\pi_{\theta, 0}}_{s, s^\prime} = \nabla_\theta \mu_\theta(s) \left. \nabla_a P(s^\prime |s, a) \right|_{a = \mu_\theta(s)}$. Indeed, let $\left(\sigma_n\right)_{n \in \mathbb{N}} \in {\big[0, r\big]^+}^\mathbb{N}$, $\sigma_n \downarrow 0$, then, applying the first condition of Conditions \ref{cond:regular_delta_appr} with $f: a \mapsto \nabla_a P(s^\prime | s, a)$ belonging to $\mathcal{F}$, we get
\begin{align*}
    &\left \lVert \nabla_\theta P^{\pi_{\theta, \sigma_n}}_{s, s^\prime} - \nabla_\theta P^{\mu_\theta}_{s, s^\prime}\right\lVert 
    \\
    &= \left\lVert \nabla_\theta \mu_\theta(s) \right\lVert_\textnormal{op} \left\lVert \int_{\mathcal{C}_{\mu_\theta(s)}} \nu_{\sigma_n}(\mu_\theta(s), a) \nabla_a P(s^\prime |s, a) \textnormal{d}a - \left. \nabla_a P(s^\prime |s, a) \right|_{a = \mu_\theta(s)} \right\lVert \underset{n \rightarrow \infty}{\longrightarrow} 0.
\end{align*}
Since $d^{\pi_{\theta, \sigma}} = \frac{1}{\det\left(A^{\pi_{\theta, \sigma}} \right)}\textnormal{adj}\left(A^{\pi_{\theta, \sigma}}\right)^\top a$ with $|\det\left(A^{\pi_{\theta, \sigma}}\right)| > 0$ for all $\sigma \in \big[0, r\big]$ and since entries of $\textnormal{adj}\left(A^{\pi_{\theta, \sigma}}\right)$ and $\det\left(A^{\pi_{\theta, \sigma}}\right)$ are polynomial functions of the entries of $P^{\pi_{\theta, \sigma}}$, it follows that $\sigma \mapsto \nabla_\theta d^{\pi_{\theta, \sigma}}$ is properly defined on $\big[0, r \big]$ and is continuous at 0, which concludes the proof of Lemma \ref{stationary_distrib}.
\end{proof}

We now proceed to prove Theorem~\ref{thm:limit_stoch_grad_technical}.

Let $\theta \in \Theta$, $\pi_\theta$ as in Theorem \ref{thm:limit_stoch_grad}, and $r > 0$ such that $\sigma \mapsto d^{\pi_{\theta, \sigma}}$, $\sigma \mapsto \nabla_\theta d^{\pi_{\theta, \sigma}}$ are well defined on $\big[0, r\big]$ and are continuous at 0. Then, the following two functions
\begin{align*}
    &\sigma \mapsto J_{\pi_{\theta, \sigma}}({\pi_{\theta, \sigma}}) = \sum_{s \in \mathcal{S}} d^{\pi_{\theta, \sigma}}(s) \int_\mathcal{A}{\pi_{\theta, \sigma}}(a|s) \bar{R}(s,a) \textnormal{d}a,\\
    &\sigma \mapsto J_{\pi_{\theta, \sigma}}(\mu_\theta) = \sum_{s \in \mathcal{S}} d^{\pi_{\theta, \sigma}}(s) \bar{R}(s, \mu_\theta(s)),
\end{align*}
are properly defined on $\big[0, r\big]$ (with $J_{\pi_{\theta, 0}}(\pi_{\theta, 0}) = J_{\pi_{\theta, 0}}(\mu_\theta) = J_{\mu_\theta}(\mu_\theta)$). Let $s \in \mathcal{S}$, by taking similar arguments as in the proof of Lemma \ref{stationary_distrib}, we have
\begin{align*}
    \nabla_\theta \int_{\mathcal{A}} \pi_{\theta, \sigma}(a |s) \bar{R}(s, a) \textnormal{d}a &= \int_{\mathcal{A}} \nabla_\theta \pi_{\theta, \sigma}(a,s) \bar{R}(s,a) \textnormal{d}a,\\
    &= \nabla_\theta \mu_\theta(s) \int_{\mathcal{C}_{\mu_\theta(s)}} \nu_\sigma(\mu_\theta(s), a) \nabla_a \bar{R}(s,a) \textnormal{d}a.
\end{align*}

Thus, $\sigma \mapsto \nabla_\theta J_{\pi_{\theta, \sigma}}(\pi_{\theta, \sigma})$ is properly defined on $\big[0, r\big]$ and
\begin{align*}
    \nabla_\theta J_{\pi_{\theta, \sigma}}(\pi_{\theta, \sigma}) &= \sum_{s \in \mathcal{S}} \nabla_\theta d^{\pi_{\theta, \sigma}}(s) \int_{\mathcal{A}} \pi_{\theta, \sigma}(a|s) \bar{R}(s,a) \textnormal{d}a 
    \\
    &\quad + \sum_{s \in \mathcal{S}} d^{\pi_{\theta, \sigma}}(s)  \nabla_\theta \int_{\mathcal{A}} \pi_{\theta, \sigma}(a|s) \bar{R}(s,a) \textnormal{d}a
    \\
    &= \sum_{s \in \mathcal{S}} \nabla_\theta d^{\pi_{\theta, \sigma}}(s) \int_{\mathcal{A}} \nu_\sigma(\mu_\theta(s), a) \bar{R}(s,a) \textnormal{d}a 
    \\
    &\quad + \sum_{s \in \mathcal{S}} d^{\pi_{\theta, \sigma}}(s) \nabla_\theta \mu_\theta(s) \int_{\mathcal{C}_{\mu_\theta(s)}} \nu_\sigma(\mu_\theta(s), a) \nabla_a \bar{R}(s,a) \textnormal{d}a.
\end{align*}

Similarly, $\sigma \mapsto \nabla_\theta J_{\pi_{\theta, \sigma}}(\mu_\theta)$ is properly defined on $\big[0, r\big]$ and
\begin{equation*}
    \nabla_\theta J_{\pi_{\theta, \sigma}}(\mu_\theta) = \sum_{s \in \mathcal{S}} \nabla_\theta d^{\pi_{\theta, \sigma}}(s) \bar{R}(s,\mu_\theta(s)) + \sum_{s \in \mathcal{S}} d^{\pi_{\theta, \sigma}}(s) \nabla_\theta \mu_\theta(s)\left. \nabla_a \bar{R}(s,a)\right|_{a = \mu_\theta(s)}
\end{equation*}

To prove continuity at $0$ of both $\sigma \mapsto \nabla_\theta J_{\pi_{\theta, \sigma}}(\pi_{\theta, \sigma})$ and $\sigma \mapsto \nabla_\theta J_{\pi_{\theta, \sigma}}(\mu_\theta)$ (with $\nabla_\theta J_{\pi_{\theta, 0}}(\pi_{\theta, 0}) = \nabla_\theta J_{\pi_{\theta, 0}}(\mu_\theta) = \nabla_\theta J_{\mu_\theta}(\mu_\theta)$), let $\pa{\sigma_n}_{n\geq 0} \downarrow 0$:

\begin{align}
    &\left\lVert\nabla_\theta J_{\pi_{\theta, \sigma_n}}(\pi_{\theta, \sigma_n}) - \nabla_\theta J_{\pi_{\theta, 0}}(\pi_{\theta,0}) \right\lVert 
    \nonumber \\
    &\leq \left\lVert\nabla_\theta J_{\pi_{\theta, \sigma_n}}(\pi_{\theta, \sigma_n}) - \nabla_\theta J_{\pi_{\theta, \sigma_n}}(\mu_\theta) \right\lVert + \left\lVert\nabla_\theta J_{\pi_{\theta, \sigma_n}}(\mu_\theta) - \nabla_\theta J_{\mu_\theta}(\mu_\theta)\right\lVert.
     \label{continuity_J}
\end{align}

For the first term of the r.h.s we have

\begin{align*}
    &\left\lVert \nabla_\theta J_{\pi_{\theta, \sigma_n}}(\pi_{\theta, \sigma_n}) - \nabla_\theta J_{\pi_{\theta, \sigma_n}}(\mu_\theta) \right\lVert \\
    &\leq \sum_{s \in \mathcal{S}} \lVert\nabla_\theta d^{\pi_{\theta, \sigma_n}}(s)\lVert \left|\int_{\mathcal{A}} \nu_{\sigma_n}(\mu_\theta(s),a) \bar{R}(s,a)\textnormal{d}a - \bar{R}(s, \mu_\theta(s))\right|\\
    & \quad + \sum_{s \in \mathcal{S}} d^{\pi_{\theta, \sigma_n}}(s) \lVert\nabla_\theta \mu_\theta(s)\lVert_\textnormal{op} \left\lVert\int_{\mathcal{A}} \nu_{\sigma_n}(\mu_\theta(s),a) \nabla_a \bar{R}(s,a)\textnormal{d}a - \left. \nabla_a \bar{R}(s, a)\right|_{a=\mu_\theta(s)}\right\lVert.
\end{align*}

Applying the first assumption in Condition \ref{cond:regular_delta_appr} with $f: a \mapsto \bar{R}(s, a)$ and $f: a \mapsto \nabla_a \bar{R}(s, a)$ belonging to $\mathcal{F}$ 
we have, for each $s \in \mathcal{S}$:
\begin{align*}
    \left|\int_{\mathcal{A}} \nu_{\sigma_n}(\mu_\theta(s),a) \bar{R}(s,a)\textnormal{d}a - \bar{R}(s, \mu_\theta(s))\right| &\underset{n \to \infty}{\longrightarrow} 0 \quad \textnormal{and}\\
    \left\lVert\int_{\mathcal{A}} \nu_{\sigma_n}(\mu_\theta(s),a) \nabla_a \bar{R}(s,a)\textnormal{d}a - \left. \nabla_a \bar{R}(s, a)\right|_{a=\mu_\theta(s)}\right\lVert &\underset{n \to \infty}{\longrightarrow} 0.
\end{align*}
Moreover, for each $s \in \mathcal{S}$, $d^{\pi_{\theta, \sigma_n}}(s) \underset{n \to \infty}{\longrightarrow} d^{\mu_\theta}(s)$ and $\nabla_\theta d^{\pi_{\theta, \sigma_n}}(s) \underset{n \to \infty}{\longrightarrow} \nabla_\theta d^{\mu_\theta}(s)$ (by Lemma \ref{stationary_distrib}), and $\lVert \nabla_\theta \mu_\theta(s) \lVert_\textnormal{op} < \infty$, 
so
\begin{equation*}
    \left\lVert \nabla_\theta J_{\pi_{\theta, \sigma_n}}(\pi_{\theta, \sigma_n}) - \nabla_\theta J_{\pi_{\theta, \sigma_n}}(\mu_\theta) \right\lVert \underset{n \to \infty}{\longrightarrow} 0.
\end{equation*}

For the second term of the r.h.s of (\ref{continuity_J}), we have
\begin{align*}
    \left\lVert\nabla_\theta J_{\pi_{\theta, \sigma_n}}(\mu_\theta) - \nabla_\theta J_{\mu_\theta}(\mu_\theta)\right\lVert &\leq \sum_{s \in \mathcal{S}} \left\lVert\nabla_\theta d^{\pi_{\theta, \sigma_n}}(s) - \nabla_\theta d^{\mu_\theta}(s)\right\lVert \left|\bar{R}(s,\mu_\theta(s))\right|\\
    & \quad + \sum_{s \in \mathcal{S}}\left|d^{\pi_{\theta, \sigma_n}}(s) - d^{\mu_\theta}(s)\right| \left\lVert \nabla_\theta \mu_\theta(s)\right\lVert_\textnormal{op} \left\lVert \left. \nabla_a \bar{R}(s,a)\right|_{a = \mu_\theta(s)} \right\lVert.
\end{align*}
Continuity at 0 of $\sigma \mapsto d^{\pi_{\theta,\sigma}}(s)$ and $\sigma \mapsto \nabla_\theta d^{\pi_{\theta,\sigma}}(s)$ for each $s \in \mathcal{S}$, boundedness of $\bar{R}(s, \cdot)$, $\nabla_a \bar{R}(s, \cdot)$ and $\nabla_\theta(s) \mu_\theta(s)$ implies that
\begin{equation*}
    \left\lVert\nabla_\theta J_{\pi_{\theta, \sigma_n}}(\mu_\theta) - \nabla_\theta J_{\mu_\theta}(\mu_\theta)\right\lVert \underset{n \to \infty}{\longrightarrow} 0.
\end{equation*}
Hence,
\begin{equation*}
    \left\lVert\nabla_\theta J_{\pi_{\theta, \sigma_n}}(\pi_{\theta, \sigma_n}) - \nabla_\theta J_{\pi_{\theta, 0}}(\pi_{\theta,0}) \right\lVert  \underset{n \to \infty}{\longrightarrow} 0.
\end{equation*}

So, $\sigma \mapsto \nabla_\theta J_{\pi_{\theta, \sigma}}(\pi_{\theta, \sigma})$ and $\nabla_\theta J_{\pi_{\theta, \sigma}}(\mu_\theta)$ are continuous at $0$:
\begin{equation*}
 \underset{\sigma \downarrow 0}{\textnormal{lim }} \nabla_\theta J_{\pi_{\theta, \sigma}} (\pi_{\theta, \sigma}) = \underset{\sigma \downarrow 0}{\textnormal{lim }} \nabla_\theta J_{\pi_{\theta, \sigma}}(\mu_\theta) = \nabla_\theta J_{\mu_\theta}(\mu_\theta).
\end{equation*}


\subsection*{Proof of Theorem \ref{thm:on-policy-AC-critic-cv}}

We will use the two-time-scale stochastic approximation analysis \cite{}. We let the policy parameter $\theta_t$ fixed as $\theta_t \equiv \theta$ when analysing the convergence of the critic step. Thus we can show the convergence of $\omega_t$ towards an $\omega_\theta$ depending on $\theta$, which will then be used to prove the convergence for the slow time-scale.

\begin{lem} \label{Boundedness_omega}
Under Assumptions 
\ref{Bound_theta}~--~\ref{Two_timesteps}, the sequence ${\omega_t^i}$ generated from (\ref{alg:On_Policy_AC_critic_step}) is bounded a.s., i.e., $\textnormal{sup}_t \lVert \omega^i_t \lVert < \infty$ a.s., for any $i \in \mathcal{N}$.
\end{lem}

The proof follows the same steps as that of Lemma B.1 in the PMLR version of \cite{FDMARL}.

\begin{lem} \label{Boundedness_J_hat}
Under Assumption
\ref{Two_timesteps}, the sequence $\{\hat{J}^i_t\}$ generated as in \ref{alg:On_Policy_AC_critic_step} is bounded a.s, i.e., $\textnormal{sup}_t \vert \hat{J}^i_t \vert < \infty$ a.s., for any $i \in \mathcal{N}$.
\end{lem}

The proof follows the same steps as that of Lemma B.2 in the PMLR version of \cite{FDMARL}.

The desired result holds since \textbf{Step 1} and \textbf{Step 2} of the proof of Theorem 4.6 in \cite{FDMARL} can both be repeated in the setting of deterministic policies.

\subsection*{Proof of Theorem \ref{thm:on-policy-AC-actor-cv}}


Let $\mathcal{F}_{t,2} = \sigma (\theta_\tau, s_\tau, \tau \leq t)$ a filtration. In addition, we define
\begin{alignat*}{3}
H(\theta, s, \omega) 
&= \nabla_{\theta} \mu_\theta(s) \cdot \nablaV{a}{Q_{\omega}(s, a)}{\mu_{\theta}(s)},
\\
H(\theta, s) 
&= H(\theta, s, \omega_\theta),
\\
h(\theta) 
&= \espL{H(\theta,s)}{s \sim d^\theta}.
\end{alignat*}

Then, for each $\theta \in \Theta$, we can introduce $\nu_\theta : \mc{S} \to \mathbb{R}^n$ the solution to the Poisson equation:
\begin{equation*}
    \pa{I - P^\theta}\nu_\theta(\cdot) = H(\theta, \cdot) - h(\theta)
\end{equation*}
that is given by $\nu_\theta(s) = \sum_{k \geq 0} \espL{H(\theta, s_k) - h(\theta) | s_0 = s}{s_{k+1} \sim P^\theta(\cdot | s_k)}$ which is properly defined (similar to the differential value function $V$).

With projection, actor update (\ref{algo1_actor_step}) becomes 
\begin{align} 
    \theta_{t+1} &= \Gamma\pab{\theta_t + \beta_{\theta, t} H(\theta_t, s_t, \omega_t)} \label{algo1_actor_update_proj}\\
    &= \Gamma\pab{\theta_t + \beta_{\theta, t} h(\theta_t) - \beta_{\theta, t} \pa{h(\theta_t) - H(\theta_t, s_t)} - \beta_{\theta, t} \pa{H(\theta_t, s_t) - H(\theta_t, s_t, \omega_t)}} \nonumber\\
    &= \Gamma\pab{\theta_t + \beta_{\theta, t} h(\theta_t) + \beta_{\theta, t} \pa{(I - P^{\theta_t})\nu_{\theta_t}(s_t)} + \beta_{\theta, t} A^1_t} \nonumber\\
    &= \Gamma\pab{\theta_t + \beta_{\theta, t} h(\theta_t) + \beta_{\theta, t} \pa{\nu_{\theta_t}(s_t) - \nu_{\theta_t}(s_{t+1})} +  \beta_{\theta, t} \pa{\nu_{\theta_t}(s_{t+1}) - P^{\theta_t}\nu_{\theta_t}(s_t)} + \beta_{\theta, t} A^1_t} \nonumber\\
    &= \Gamma\pab{\theta_t +  \beta_{\theta, t} \pa{h(\theta_t) + A^1_t + A^{2}_t + A^3_t}} \nonumber
\end{align}
where
\begin{align*}
    &A^1_t = H(\theta_t, s_t, \omega_t) - H(\theta_t, s_t),
    \\
    &A^2_t = \nu_{\theta_t}(s_t) - \nu_{\theta_t}(s_{t+1}),
    \\
    &A^3_t = \nu_{\theta_t}(s_{t+1}) - P^{\theta_t}\nu_{\theta_t}(s_t).
\end{align*}

For $r < t$ we have
\begin{align*}
    \sum_{k=r}^{t-1} \beta_{\theta,k} A^2_k &= \sum_{k=r}^{t-1} \beta_{\theta,k} \pa{\nu_{\theta_k}(s_k) - \nu_{\theta_k}(s_{k+1})}\\
    &= \sum_{k=r}^{t-1} \beta_{\theta,k} \pa{\nu_{\theta_k}(s_k) - \nu_{\theta_{k+1}}(s_{k+1})} + \sum_{k=r}^{t-1} \beta_{\theta,k} \pa{\nu_{\theta_{k+1}}(s_{k+1}) - \nu_{\theta_{k}}(s_{k+1})}\\
    &= \sum_{k=r}^{t-1} \pa{\beta_{\theta,k+1} - \beta_{\theta, k}} \nu_{\theta_{k+1}}(s_{k+1}) + \beta_{\theta_r} \nu_{\theta_r} (s_r) - \beta_{\theta_t} \nu_{\theta_t} (s_t) + \sum_{k=r}^{t-1} \epsilon^{(2)}_k  \\
    &= \sum_{k=r}^{t-1} \epsilon^{(1)}_k + \sum_{k=r}^{t-1} \epsilon^{(2)}_k + \eta_{r,t}
\end{align*}
where
\begin{align*}
    &\epsilon^{(1)}_k = \pa{\beta_{\theta,k+1} - \beta_{\theta, k}} \nu_{\theta_{k+1}}(s_{k+1}),\\
    &\epsilon^{(2)}_k = \beta_{\theta,k} \pa{\nu_{\theta_{k+1}}(s_{k+1}) - \nu_{\theta_{k}}(s_{k+1})},\\
    &\eta_{r,t} = \beta_{\theta_r} \nu_{\theta_r} (s_r) - \beta_{\theta_t} \nu_{\theta_t} (s_t).
\end{align*}

\begin{lem} \label{lem:A2cv}
$\sum_{k=0}^{t-1} \beta_{\theta,k} A_k^2$ converges a.s. for $t \to \infty$
\end{lem}

\begin{proof}[Proof of Lemma \ref{lem:A2cv}]
Since $\nu_\theta(s)$ is uniformly bounded for $\theta \in \Theta, s \in \mc{S}$, we have for some $K > 0$
\begin{equation*}
    \sum_{k=0}^{t-1} \palV{\epsilon_k^{(1)}}
    \leq K \sum_{k=0}^{t-1} \pav{\beta_{\theta,k+1} - \beta_{\theta, k}}
\end{equation*}
which converges given Assumption \ref{Two_timesteps}.

Moreover, since $\mu_\theta(s)$ is twice continuously differentiable, $\theta \mapsto \nu_{\theta}(s)$ is Lipschitz for each $s$, and so we have
\begin{align*}
    \sum_{k=0}^{t-1} \palV{\epsilon_k^{(2)}}
    &\leq \sum_{k=0}^{t-1} \beta_{\theta, k} \palV{\nu_{\theta_k}(s_{k+1}) - \nu_{\theta_{k+1}}(s_{k+1})}\\
    &\leq K^2 \sum_{k=0}^{t-1} \beta_{\theta, k} \palV{\theta_k - \theta_{k+1}}\\
    &\leq K^3 \sum_{k=0}^{t-1} \beta_{\theta, k}^2.
\end{align*}
Finally, $\limit{t \to \infty} \palV{\eta_{0,t}} = \beta_{\theta,0} \palV{\nu_{\theta_0}(s_0)} < \infty$ a.s.


Thus, $\sum_{k=0}^{t-1} \palV{\beta_{\theta,k} A_k^2} \leq \sum_{k=0}^{t-1} \palV{\epsilon_k^{(1)}} + \sum_{k=0}^{t-1} \palV{\epsilon_k^{(2)}} + \palV{\eta_{0,t}}$ converges a.s.
\end{proof}

\begin{lem} \label{lem:A3cv}
$\sum_{k=0}^{t-1} \beta_{\theta,k} A_k^3$ converges a.s. for $t \to \infty$.
\end{lem}

\begin{proof}[Proof of Lemma \ref{lem:A3cv}]
We set 
\begin{equation*}
    Z_t = \sum_{k=0}^{t-1} \beta_{\theta,k} A_k^3 = \sum_{k=0}^{t-1} \beta_{\theta,k} \pa{\nu_{\theta_k}(s_{k+1}) - P^{\theta_k}\nu_{\theta_k}(s_k)}.
\end{equation*}
Since $Z_t$ is $\mc{F}_t$-adapted and $\esp{\nu_{\theta_t}(s_{t+1})|\mc{F}_t} = P^{\theta_t}\nu_{\theta_t}(s_t)$, $Z_t$ is a martingale. The remaining of the proof is now similar to the proof of Lemma 2 on page 224 of \cite{Benveniste:1990:adaptive_algorithm_stochastic_approximation}.
\end{proof}

Let $g^i(\theta_t) = \mathbb{E}_{s_t \sim d^{\theta_t}} \big[\psi^i_t \cdot \xi^i_{t, \theta_t} \vert \mathcal{F}_{t,2} \big]$ and $g(\theta) = \big[g^1(\theta), \dots, g^N (\theta) \big]$. We have
$$g^i(\theta_t) = \sum_{s_t \in \mathcal{S}} d^{\theta_t}(s_t) \cdot \psi^i_t \cdot \xi^i_{t, \theta_t}.$$

Given 
(\ref{omega_expr}), $\theta \mapsto \omega_\theta$ is continuously differentiable and $\theta \mapsto \nabla_\theta \omega_\theta$ is bounded so $\theta \mapsto \omega_\theta$ is Lipschitz-continuous. Thus $\theta \mapsto \xi^i_{t,\theta}$ is Lipschitz-continuous for each $s_t \in \mathcal{S}$. 
Due to our regularity assumptions, 
$\theta \mapsto \psi_{t, \theta_t}^i$ is also continuous for each $i \in \mathcal{N}, s_t \in \mathcal{S}$. Moreover, $\theta \mapsto d^\theta(s)$ is also Lipschitz continuous for each $s\in\mathcal{S}$. Hence, $\theta \mapsto g(\theta)$ is Lipschitz-continuous in $\theta$ and the ODE (\ref{ODE_actor}) is well-posed. This holds even when using compatible features.

By critic faster convergence, we have $\textnormal{lim}_{t \to \infty} \lVert \xi_t^i - \xi_{t, \theta_t}^i \lVert = 0$ so $\textnormal{lim}_{t \to \infty} A^1_t = 0$. 

Hence, by Kushner-Clark lemma \cite{Kushner_Clark} (pp 191-196) we have that the update in (\ref{algo1_actor_update_proj}) converges a.s. to the set of asymptotically stable equilibria of the ODE (\ref{ODE_actor}).

\subsection*{Proof of Theorem \ref{thm:off-policy-AC-critic-cv}}


We use the two-time scale technique: since critic updates at a faster rate than the actor, we let the policy parameter $\theta_t$ to be fixed as $\theta$ when analysing the convergence of the critic update.

\begin{lem} \label{lem_bounded_lambda}
Under Assumptions 
\ref{asm_random_matrix},  \ref{asm_compatible_features_R} and \ref{asm_algo2_two_timesteps}, for any $i \in \mathcal{N}$, sequence $\{\lambda_t^i\}$ generated from (\ref{alg:Off_Policy_AC_critic_step}) is bounded almost surely.
\end{lem}

To prove this lemma we verify the conditions for \textbf{Theorem A.2} of \cite{FDMARL} to hold. We use $\{\mathcal{F}_{t,1}\}$ to denote the filtration with $\mathcal{F}_{t,1} = \sigma(s_\tau, C_{\tau-1}, a_{\tau-1}, r_\tau, \lambda_\tau, \tau \leq t)$. With $\lambda_t = \big[(\lambda^1_t)^\top, \dots, (\lambda^N_t)^\top\big]^\top$, critic step (\ref{alg:Off_Policy_AC_critic_step}) has the form:
\begin{equation} \label{eq_algo2_rec_lambda}
    \lambda_{t+1} = (C_t \otimes I)\left(\lambda_t + \beta_{\lambda, t}\cdot y_{t+1}\right)
\end{equation}
with $y_{t+1} = \left(\delta_t^1w(s_t,a_t)^\top, \dots, \delta_t^N w(s_t, a_t)^\top\right)^\top \in \mathbb{R}^{KN}$, $\otimes$ denotes Kronecker product and $I$ is the identity matrix. Using the same notation as in \textbf{Assumption A.1} from \cite{FDMARL}, we have:
\begin{align*}
    &h^i(\lambda_t^i,s_t) = \mathbb{E}_{a \sim \pi}\big[\delta_t^i w(s_t,a)^\top \vert \mathcal{F}_{t,1}\big] = \int_\mathcal{A}\pi(a|s_t)(R^i(s_t,a) -  w(s_t,a) \cdot \lambda_t^i) w(s_t,a)^\top \textnormal{d}a,\\
    &M^i_{t+1} = \delta_t^i w(s_t,a_t)^\top - \mathbb{E}_{a \sim \pi}\big[\delta_t^i w(s_t,a)^\top \vert \mathcal{F}_{t,1}\big],\\
    &\bar{h}^i(\lambda_t) = A_{\pi, \theta}^i \cdot d_\pi^s - B_{\pi, \theta} \cdot \lambda_t, \qquad \textnormal{where } A_{\pi,\theta}^i = \pab{\int_\mathcal{A}\pi(a|s) R^i(s,a)  w(s,a)^\top \textnormal{d}a, s \in \mathcal{S}}.
\end{align*}
Since feature vectors are uniformly bounded for any $s \in \mathcal{S}$ and $a \in \mathcal{A}$, $h^i$ is Lipschitz continuous in its first argument. Since, for $i \in \mathcal{N}$, the $r^i$ are also uniformly bounded, $\mathbb{E}\big[\lVert M_{t+1}\lVert^2 | \mathcal{F}_{t,1}\big] \leq K \cdot (1 + \lVert\lambda_t\lVert^2)$ for some $K > 0$. Furthermore, finiteness of $|\mathcal{S}|$ ensures that, a.s., $\lVert \bar{h}(\lambda_t) - h(\lambda_t, s_t) \lVert^2 \leq K^\prime \cdot (1 + \lVert \lambda_t \lVert^2)$. Finally, $h_\infty(y)$ exists and has the form
\begin{equation*}
    h_\infty(y) = - B_{\pi, \theta} \cdot y.
\end{equation*}
From Assumption \ref{asm_compatible_features_R}, we have that $-B_{\pi, \theta}$ is a Hurwitcz matrix, thus the origin is a globally asymptotically stable attractor of the ODE $\dot{y} = h_\infty(y)$. Hence \textbf{Theorem A.2} of \cite{FDMARL} applies, which concludes the proof of Lemma \ref{lem_bounded_lambda}.
\bigbreak

We introduce the following operators as in \cite{FDMARL}:
\begin{itemize}
    \item $\langle\cdot\rangle: \mathbb{R}^{KN} \to \mathbb{R}^K$
    \begin{equation*}
        \langle\lambda\rangle = \frac{1}{N}(\textbf{1}^\top \otimes I)\lambda = \frac{1}{N}\sum_{i \in \mathcal{N}} \lambda^i.
    \end{equation*}
    \item $\mathcal{J} = \left(\frac{1}{N}\textbf{1}\textbf{1}^\top\otimes I\right): \mathbb{R}^{KN} \to \mathbb{R}^{KN}$ such that $\mathcal{J}\lambda = \textbf{1} \otimes \langle\lambda\rangle$.
    \item $\mathcal{J}_\bot = I - \mathcal{J}: \mathbb{R}^{KN} \to \mathbb{R}^{KN}$ and we note $\lambda_\bot = \mathcal{J}_\bot \lambda = \lambda - \textbf{1} \otimes \langle\lambda\rangle$.
\end{itemize}

We then proceed in two steps as in \cite{FDMARL}, firstly by showing the convergence a.s. of the disagreement vector sequence $\{\lambda_{\bot, t}\}$ to zero, secondly showing that the consensus vector sequence $\{\langle\lambda_{t}\rangle\}$ converges to the equilibrium such that $\langle\lambda_t\rangle$ is solution to (\ref{eq_algo2_consensus_cv}).

\begin{lem}
Under Assumptions 
\ref{asm_random_matrix},  \ref{asm_compatible_features_R} and \ref{asm_algo2_two_timesteps}, for any $M > 0$, we have
\begin{equation*}
    \underset{t}{\textnormal{sup }} \mathbb{E}\Big[\lVert\beta_{\lambda, t}^{-1}\lambda_{\bot,t}\lVert^2 \mathbbm{1}_{\{\sup_t\lVert\lambda_t\lVert\leq M\}}\Big] < \infty.
\end{equation*}
\end{lem}
Since dynamic of $\{\lambda_t\}$ described by (\ref{eq_algo2_rec_lambda}) is similar to (5.2) in \cite{FDMARL} we have
\begin{equation} \label{eq_algo2_rec_lamb}
    \mathbb{E}\Big[\lVert\beta_{\lambda, t+1}^{-1}\lambda_{\bot,t+1}\lVert^2|\mathcal{F}_{t,1}\Big] = \frac{\beta_{\lambda,t}^2}{\beta_{\lambda,t+1}^2} \rho \left(\lVert\beta_{\lambda, t}^{-1}\lambda_{\bot,t}\lVert^2 + 2 \cdot \lVert\beta_{\lambda, t}^{-1}\lambda_{\bot,t}\lVert \cdot \mathbb{E}(\lVert y_{t+1}\lVert^2|\mathcal{F}_{t,1})^{\frac{1}{2}} + \mathbb{E}(\lVert y_{t+1}\lVert^2|\mathcal{F}_{t,1})\right)
\end{equation}
where $\rho$ represents the spectral norm of $\mathbb{E}\big[C_t^\top \cdot (I - \textbf{1} \textbf{1}^\top / N) \cdot C_t \big]$, with $\rho \in \left[0, 1\right)$ by Assumption \ref{asm_random_matrix}. Since $y_{t+1}^i = \delta_t^i \cdot  w(s_t,a_t)^\top$ we have
\begin{align*}
    \mathbb{E}\Big[\lVert y_{t+1} \lVert^2|\mathcal{F}_{t,1}\Big] &= \mathbb{E}\Big[\sum_{i \in \mathcal{N}} \lVert(r^i(s_t,a_t) -  w(s_t,a_t)\lambda_t^i)\cdot  w(s_t,a_t)^\top\lVert^2 \vert \mathcal{F}_{t,1}\Big]\\
    &\leq 2 \cdot \mathbb{E}\Big[\sum_{i \in \mathcal{N}} \lVert r^i(s_t,a_t) w(s_t,a_t)^\top\lVert^2 + \lVert w(s_t,a_t)^\top\lVert^4 \cdot \lVert \lambda_t^i\lVert^2 \vert \mathcal{F}_{t,1}\Big].
\end{align*}
By uniform boundedness of $r(s,\cdot)$ and $ w(s,\cdot)$ 
(Assumptions 
\ref{asm_compatible_features_R}) and finiteness of $\mathcal{S}$, there exists $K_1 > 0$ such that
\begin{equation*}
    \mathbb{E}\Big[\lVert y_{t+1} \lVert^2|\mathcal{F}_{t,1}\Big] \leq K_1(1 + \lVert \lambda_t \lVert^2).
\end{equation*}
Thus, for any $M > 0$ there exists $K_2 >0$ such that, on the set $\{\sup_{\tau\leq t} \lVert\lambda_\tau\lVert < M\}$,
\begin{equation} \label{eq_y_bounded}
    \mathbb{E}\Big[\lVert y_{t+1} \lVert^2 \mathbbm{1}_{\{\sup_{\tau\leq t} \lVert\lambda_\tau\lVert < M\}}|\mathcal{F}_{t,1}\Big] \leq K_2.
\end{equation}

We let $v_t = \lVert\beta_{\lambda, t}^{-1}\lambda_{\bot,t}\lVert^2 \mathbbm{1}_{\{\sup_{\tau\leq t} \lVert\lambda_\tau\lVert < M\}}$. Taking expectation over (\ref{eq_algo2_rec_lamb}), noting that $\mathbbm{1}_{\{\sup_{\tau\leq t+1} \lVert\lambda_\tau\lVert < M\}} \leq \mathbbm{1}_{\{\sup_{\tau\leq t} \lVert\lambda_\tau\lVert < M\}}$ we get
\begin{equation*}
    \mathbb{E}(v_{t+1}) \leq \frac{\beta_{\lambda,t}^2}{\beta_{\lambda,t+1}^2} \rho \left(\mathbb{E}(v_t) + 2 \sqrt{\mathbb{E}(v_t)} \cdot \sqrt{K_2} + K_2\right)
\end{equation*}
which is the same expression as (5.10) in \cite{FDMARL}. So similar conclusions to the ones of \textbf{Step 1} of \cite{FDMARL} holds:
\begin{align}
    &\underset{t}{\textnormal{sup } } \mathbb{E}\Big[\lVert\beta_{\lambda, t}^{-1}\lambda_{\bot,t}\lVert^2 \mathbbm{1}_{\{\sup_t\lVert\lambda_t\lVert\leq M\}}\Big] < \infty \label{eq_sup_lamb}\\ 
    \textnormal{and}\qquad& \underset{t}{\textnormal{lim }} \lambda_{\bot, t} = 0 \textnormal{ a.s.}
\end{align}

We now show convergence of the consensus vector $\textbf{1} \otimes \langle \lambda_t \rangle$. Based on (\ref{eq_algo2_rec_lambda}) we have
\begin{align*}
    \langle \lambda_{t+1} \rangle &= \langle(C_t \otimes I)(\textbf{1} \otimes \langle \lambda_t \rangle + \lambda_{\bot, t} + \beta_{\lambda,t} y_{t+1})\rangle\\
    &= \langle\lambda_t\rangle + \langle\lambda_{\bot,t}\rangle + \beta_{\lambda,t} \langle(C_t \otimes I)(y_{t+1} + \beta_{\lambda,t}^{-1} \lambda_{\bot,t})\rangle\\
    &= \langle\lambda_t\rangle + \beta_{\lambda,t} (h(\lambda_t, s_t) + M_{t+1})
\end{align*}
where $h(\lambda_t, s_t) = \mathbb{E}_{a_t \sim \pi}\big[\langle y_{t+1}\rangle|\mathcal{F}_t\big]$ and $M_{t+1} = \langle(C_t \otimes I)(y_{t+1} + \beta_{\lambda,t}^{-1} \lambda_{\bot,t})\rangle - \mathbb{E}_{a_t \sim \pi}\big[\langle y_{t+1}\rangle|\mathcal{F}_t\big]$. Since $\langle\delta_t\rangle = \bar{r}(s_t,a_t) -  w(s_t,a_t)\langle\lambda_t\rangle$, we have
\begin{equation*}
    h(\lambda_t,s_t) = \mathbb{E}_{a_t\sim\pi} (\bar{r}(s_t,a_t) w(s_t,a_t)^\top|\mathcal{F}_t) + \mathbb{E}_{a_t\sim\pi}( w(s_t,a_t)\langle\lambda_t\rangle\cdot w(s_t,a_t)^\top|\mathcal{F}_{t,1})
\end{equation*}
so $h$ is Lipschitz-continuous in its first argument. Moreover, since $\langle \lambda_{\bot,t} \rangle = 0$ and $\textbf{1}^\top \mathbb{E}(C_t|\mathcal{F}_{t,1}) = \textbf{1}^\top$ a.s.:
\begin{align*}
    \mathbb{E}_{a_t \sim \pi}\big[\langle(C_t \otimes I)(y_{t+1} + \beta_{\lambda,t}^{-1} \lambda_{\bot,t})\rangle|\mathcal{F}_{t,1}\big] &= \mathbb{E}_{a_t \sim \pi}\Big[\frac{1}{N}(\textbf{1}^\top \otimes I)(C_t \otimes I)(y_{t+1} + \beta_{\lambda,t}^{-1} \lambda_{\bot,t})|\mathcal{F}_{t,1}\Big]\\
    &= \frac{1}{N}(\textbf{1}^\top \otimes I)(\mathbb{E}(C_t| \mathcal{F}_{t,1}) \otimes I)\mathbb{E}_{a_t \sim \pi}\big[ y_{t+1} + \beta_{\lambda,t}^{-1} \lambda_{\bot,t}|\mathcal{F}_{t,1}\big]\\
    &= \frac{1}{N}(\textbf{1}^\top \mathbb{E}(C_t|\mathcal{F}_{t,1})\otimes I)\mathbb{E}_{a_t \sim \pi}\big[ y_{t+1} + \beta_{\lambda,t}^{-1} \lambda_{\bot,t}|\mathcal{F}_{t,1}\big]\\
    &=\mathbb{E}_{a_t \sim \pi}\big[\langle y_{t+1}\rangle|\mathcal{F}_{t,1}\big] \textnormal{ a.s.}
\end{align*}
So $\{M_t\}$ is a martingale difference sequence. Additionally we have
\begin{equation*}
    \mathbb{E}\big[\lVert M_{t+1}\lVert^2|\mathcal{F}_{t,1}\big] \leq 2 \cdot \mathbb{E}\big[\lVert y_{t+1} + \beta^{-1}_{\lambda,t} \lambda_{\bot,t}\lVert^2_{G_t}|\mathcal{F}_{t,1}\big] + 2 \cdot \lVert \mathbb{E}\big[\langle y_{t+1} \rangle | \mathcal{F}_{t,1} \big] \lVert^2 
\end{equation*}
with $G_t = N^{-2} \cdot C_t^\top \textbf{1} \textbf{1}^\top C_t \otimes I$ whose spectral norm is bounded for $C_t$ is stochastic. From (\ref{eq_y_bounded}) and (\ref{eq_sup_lamb}) we have that, for any $M>0$, over the set $\{\sup_t \lVert\lambda_t\lVert \leq M\}$, there exists $K_3, K_4 < \infty$ such that
\begin{equation*}
    \mathbb{E}\big[\lVert y_{t+1} + \beta^{-1}_{\lambda,t} \lambda_{\bot,t}\lVert^2_{G_t}|\mathcal{F}_{t,1}\big]\mathbbm{1}_{\{\sup_t \lVert\lambda_t\lVert \leq M\}} \leq K_3 \cdot \mathbb{E}\big[\lVert y_{t+1}\lVert^2 +
    \lVert\beta^{-1}_{\lambda,t} \lambda_{\bot,t}\lVert^2|\mathcal{F}_{t,1}\big]\mathbbm{1}_{\{\sup_t \lVert\lambda_t\lVert \leq M\}} \leq K_4.
\end{equation*}
Besides, since $r^i_{t+1}$ and $ w$ are uniformly bounded, there exists $K_5 < \infty$ such that $\lVert \mathbb{E}\big[\langle y_{t+1} \rangle | \mathcal{F}_{t,1} \big] \lVert^2 \leq K_5 \cdot (1 + \lVert\langle\lambda_t\rangle\lVert^2)$. Thus, for any $M > 0$, there exists some $K_6 < \infty$ such that over the set $\{\sup_t \lVert\lambda_t\lVert \leq M\}$
\begin{equation*}
    \mathbb{E}\big[\lVert M_{t+1} \lVert^2 |\mathcal{F}_{t,1}\big] \leq K_6 \cdot (1 + \lVert \langle \lambda_t \rangle\lVert^2).
\end{equation*}
Hence, for any $M>0$, assumptions (a.1) - (a.5) of B.1. from \cite{FDMARL} are verified on the set $\{\sup_t \lVert\lambda_t\lVert \leq M\}$.
Finally, we consider the ODE asymptotically followed by $\langle\lambda_t\rangle$:
\begin{equation*}
    \dot{\langle\lambda_t\rangle} = -B_{\pi,\theta} \cdot \langle\lambda_t\rangle + A_{\pi,\theta}\cdot d^\pi
\end{equation*}
which has a single globally asymptotically stable equilibrium $\lambda^* \in \mathbb{R}^K$, since $B_{\pi,\theta}$ is positive definite: $\lambda^* = B_{\pi,\theta}^{-1} \cdot A_{\pi,\theta} \cdot d^\pi$.
By Lemma \ref{lem_bounded_lambda}, $\sup_t \lVert \langle \lambda_t \rangle \lVert < \infty$ a.s., all conditions to apply \textbf{Theorem B.2.} of \cite{FDMARL} hold a.s., which means that $ \langle \lambda_t \rangle \underset{t\to\infty}{\longrightarrow} \lambda^*$ a.s. As $\lambda_t = \textbf{1} \otimes \langle\lambda_t\rangle + \lambda_{\bot,t}$ and $\lambda_{\bot,t}\underset{t\to\infty}{\longrightarrow}0 $ a.s., we have for each $i \in \mathcal{N}$, a.s.,
\begin{equation*}
    \lambda^i_t \underset{t\to\infty}{\longrightarrow} B_{\pi,\theta}^{-1} \cdot A_{\pi,\theta} \cdot d^\pi.
\end{equation*}

\subsection*{Proof of Theorem \ref{thm:off-policy-AC-actor-cv}}

Let $\mathcal{F}_{t,2} = \sigma (\theta_\tau, \tau \leq t)$ be the $\sigma$-field generated by $\{\theta_\tau, \tau \leq t\}$, and let
\begin{alignat*}{2}
&\zeta_{t,1}^i = \psi^i_t \cdot \xi^i_t - \espL{\psi_t^i  \cdot \xi^i_t \vert \mathcal{F}_{t,2}}{s_t \sim d^\pi}, \hspace{4em}
&\zeta_{t,2}^i = \espL{\psi_t^i \cdot (\xi_t^i - \xi_{t, \theta_t}^i) \vert \mathcal{F}_{t, 2}}{s_t\sim d^\pi}.
\end{alignat*}

With local projection, actor update (\ref{algo2_actor_step}) becomes 
\begin{equation} \label{algo2_actor_update_proj}
    \theta_{t+1}^i = \Gamma^i\pab{\theta^i_t + \beta_{\theta,t} \espL{\psi_t^i \cdot \xi_{t, \theta_t}^i \vert \mathcal{F}_{t,2}}{s_t \sim d^\pi} + \beta_{\theta,t} \zeta_{t, 1}^i + \beta_{\theta,t} \zeta_{t, 2}^i}.
\end{equation}

So with $h^i(\theta_t) = \espL{\psi^i_t \cdot \xi^i_{t, \theta_t} \vert \mathcal{F}_{t,2}}{s_t \sim d^\pi}$ and $h(\theta) = \big[h^1(\theta), \dots, h^N (\theta) \big]$, we have
$$h^i(\theta_t) = \sum_{s_t \in \mathcal{S}} d^\pi(s_t) \cdot \psi^i_t \cdot \xi^i_{t, \theta_t}.$$

Given 
(\ref{omega_expr}), $\theta \mapsto \omega_\theta$ is continuously differentiable and $\theta \mapsto \nabla_\theta \omega_\theta$ is bounded so $\theta \mapsto \omega_\theta$ is Lipschitz-continuous. Thus $\theta \mapsto \xi^i_{t,\theta}$ is Lipschitz-continuous for each $s_t \in \mathcal{S}$.  
Our regularity assumptions ensure that 
$\theta \mapsto \psi_{t, \theta_t}^i$ is continuous for each $i \in \mathcal{N}, s_t \in \mathcal{S}$. Moreover, $\theta \mapsto d^\theta(s)$ is also Lipschitz continuous for each $s\in\mathcal{S}$. Hence, $\theta \mapsto g(\theta)$ is Lipschitz-continuous in $\theta$ and the ODE (\ref{ODE_actor}) is well-posed. This holds even when using compatible features.

By critic faster convergence, we have $\textnormal{lim}_{t \to \infty} \lVert \xi_t^i - \xi_{t, \theta_t}^i \lVert = 0$. 

Let $M_t^i = \sum_{\tau = 0}^{t-1} \beta_{\theta,\tau} \zeta_{\tau,1}^i$. ${M_t^i}$ is a martingale sequence with respect to $\mathcal{F}_{t,2}$.  Since $\{\omega_t\}_t, \{\nabla_a \phi_k(s,a)\}_{s, k}$, and $\{\nabla_\theta \mu_\theta(s)\}_s$ are bounded (Lemma \ref{Boundedness_omega}, Assumption \ref{Linear_appr}),
it follows that the sequence  $\paa{\zeta_{t,1}^i}$ is bounded. Thus, by Assumption \ref{Two_timesteps}, $\sum_t \esp{\palV{M_{t+1}^i - M_t^i}^2 | \mathcal{F}_{t,2}} = \sum_t \palV{\beta_{\theta,t} \zeta_{t,1}^i}^2 < \infty$ a.s. The martingale convergence theorem ensures that $\paa{M^i_t}$ converges a.s. Thus, for any $\epsilon > 0$,
\begin{equation*}
    \underset{t}{\textnormal{lim }} \mathbb{P}\pa{\underset{n \geq t}{\textnormal{sup }} \palV{\sum_{\tau = t}^n \beta_{\theta,\tau} \zeta_{\tau,1}^i} \geq \epsilon} = 0.
\end{equation*}

Hence, by Kushner-Clark lemma \cite{Kushner_Clark} (pp 191-196) we have that the update in (\ref{algo2_actor_update_proj}) converges a.s. to the set of asymptotically stable equilibria of the ODE (\ref{ODE_actor}).

